\documentclass[lettersize,journal]{IEEEtran}
\usepackage{amsmath,amsfonts,amssymb}
\usepackage{algorithmic}
\usepackage{algorithm}
\usepackage{array}
\usepackage[caption=false,font=normalsize,labelfont=sf,textfont=sf]{subfig}
\usepackage{textcomp}
\usepackage{stfloats}
\usepackage{flushend}
\usepackage{url}
\usepackage{verbatim}
\usepackage{graphicx}
\usepackage{cite}
\usepackage{multirow}
\usepackage{makecell}
\usepackage{booktabs}

\newtheorem{myDef}{Definition}

\newcommand{\eg}{\emph{e.g., }}
\newcommand{\ie}{\emph{i.e., }}

\newcommand{\eat}[1]{}
\ifodd 0

\newcommand{\TODO}[1]{{\color{red}TODO:{#1}}}
\newcommand\beftext[1]{{\color[rgb]{0.5,0.5,0.5}{BEFORE:#1}}}
\else

\newcommand{\TODO}[1]{}
\newcommand{\beftext}[1]{}
\fi

\begin{document}

\title{Atmospheric Diffusion-Guided Spatio-Temporal Transformer for Nuclear Radiation Forecasting}

\author{Tengfei Lyu, Jindong Han, and Hao Liu

\thanks{Tengfei Lyu is with the Thrust of Artificial Intelligence, The
Hong Kong University of Science and Technology (Guangzhou), Guangzhou, China, and The Hong Kong University of Science and Technology, Hong Kong, SAR, China
(e-mail: tlyu077@connect.hkust-gz.edu.cn).}

\thanks{Jindong Han is with the School of Artificial Intelligence, Shandong University, Jinan, Shandong, China (e-mail: jindong.han@sdu.edu.cn).}

\thanks{Hao Liu is with the Thrust of Artificial Intelligence, The
Hong Kong University of Science and Technology (Guangzhou), Guangzhou, China, and also with the Department of Computer Science and Engineering,  The Hong Kong University of Science and Technology, Hong Kong, SAR, China (e-mail: liuh@ust.hk).}

\thanks{Corresponding authors: Hao Liu.}

% <-this % stops a space
}

\markboth{IEEE Transactions on Knowledge and Data Engineering}%
{Lyu \MakeLowercase{\textit{et al.}}: Atmospheric Diffusion-Guided Spatio-Temporal Transformer for Nationwide Nuclear Radiation Forecasting}

% \IEEEpubid{0000--0000/00\$00.00~\copyright~2026 IEEE}

\maketitle

\begin{abstract}
Nuclear radiation, the energy released during atomic decay, poses persistent risks to public health and the environment, and concerns have only grown since the Fukushima accident and the recent commencement of treated-water discharge. 
Modern monitoring networks now record radiation levels and accompanying weather conditions at thousands of stations, opening the door to nationwide forecasting that can inform emergency response, agricultural advisories, and routine public-safety decisions. 
However, turning this abundance of monitoring data into reliable forecasts is difficult for three reasons. 
First, the time series at each station are highly non-stationary, shaped by radioactive decay, weather variability, and irregular human interventions. 
Second, monitoring stations are severely unevenly distributed in space. Roughly 78\% of Japan's stations sit in less than 6\% of the country, clustered near Fukushima, which breaks the assumptions of standard graph-based models. 
Third, radiation co-evolves with heterogeneous context such as wind, temperature, and humidity through atmospheric transport processes that purely data-driven models struggle to capture from observations alone.
In this study, we introduce \emph{NRFormer+}, a spatio-temporal Transformer for nationwide nuclear radiation forecasting. NRFormer+ couples non-stationary temporal attention and density-adaptive spatial attention with a new atmospheric diffusion module that estimates how meteorology drives radiation dispersion and injects this physical signal into the network as an architectural prior. 
We curate two complementary benchmarks, \emph{Japan-4H} and \emph{Japan-1D}, covering 3{,}627 radiation and 228 meteorological stations over four years. NRFormer+ delivers state-of-the-art accuracy on both datasets across all $13$ baselines, reducing sudden-change MAE by up to 19.1\% over the strongest baseline at comparable inference latency. Our code and datasets are publicly available at \url{https://github.com/tfeilyu/NRFormer_Plus}.
\end{abstract}

\section{Introduction}
\label{Introduction}

Nuclear power has supplied clean, dispatchable electricity for more than half a century, but it carries the low-probability, high-consequence risk of radioactive release. The 1986 Chernobyl accident dispersed radionuclides across the European continent~\cite{ager2019wildfire}, and the 2011 Fukushima Daiichi accident contaminated soil, marine ecosystems, and the food chain over thousands of square kilometres of northeastern Japan~\cite{steinhauser2014comparison}. Because the principal long-lived contaminant has a physical half-life of roughly 30 years, the radioecological footprint of such events persists across generations, and the recent commencement of treated-water discharge from the Fukushima site has renewed public and regulatory attention on long-term radiological monitoring. In response, many countries now operate dense networks of monitoring stations. Yet raw measurements alone are not enough. Governments and citizens increasingly demand \emph{forecasts} of future radiation levels to support evacuation planning, agricultural advisories, and pre-positioning of protective equipment, where even a few hours of additional lead time materially reduces socio-economic loss.

\begin{figure}[!t]
  \centering
  \includegraphics[width=1\linewidth]{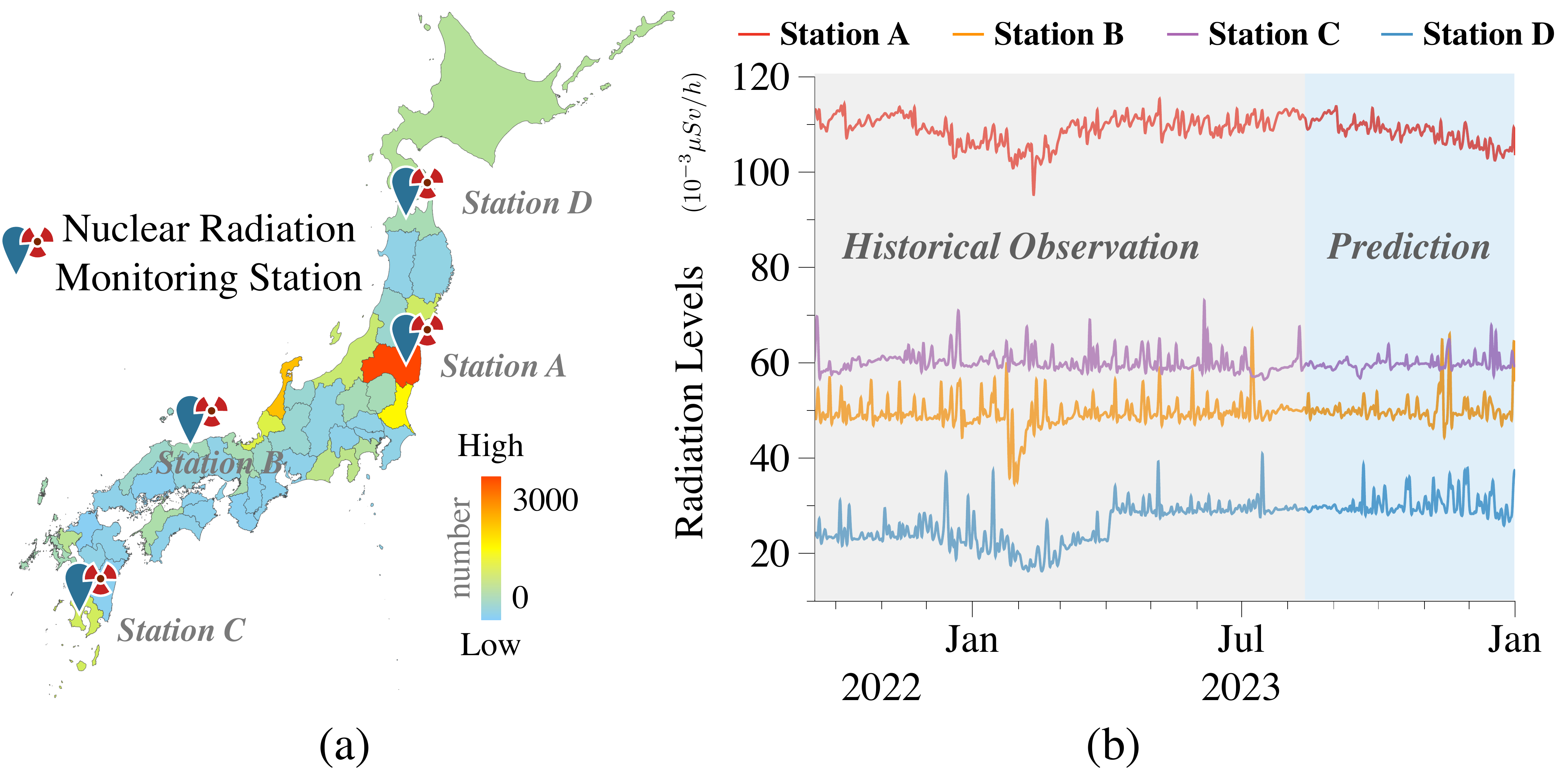}
  \caption{Nationwide nuclear radiation forecasting on Japan. (a)~Spatial distribution of monitoring stations; high-risk areas (\ie those near nuclear power plants) host disproportionately more stations. (b)~Time-varying radiation levels at representative stations.}
  \label{fig:problem-define}
\end{figure}

In this paper, we study \emph{nationwide nuclear radiation forecasting}, the task of producing station-level multi-step predictions across an entire country from a heterogeneous mix of radiation, meteorological, and geographic observations. Figure~\ref{fig:problem-define} illustrates the setting in Japan, where thousands of gamma-radiation monitors are deployed with density determined by proximity to nuclear facilities rather than by uniform geographic coverage. The dispersion of radiation between stations is governed by atmospheric circulation~\cite{qiao2011predicting,stockie2011mathematics}, in which wind transports radionuclides across regions while temperature and humidity modulate mixing, deposition, and removal.

Existing approaches lie largely outside this regime. Operational atmospheric-dispersion solvers such as HYSPLIT~\cite{stein2015hysplit} and FLEXPART~\cite{stohl2005flexpart} simulate radiation transport mechanistically. In forward mode they presuppose a specified emission source, and although their backward and data-assimilation variants can reconstruct unknown sources from sparse observations~\cite{stohl2012xenon}, these variants are designed for hindcasting discrete release events and in either mode still require dense meteorological fields and substantial compute, which makes them ill-suited to continuous nationwide station-level forecasting. Conversely, generic Spatio-Temporal Graph Neural Networks~\cite{wu2019graph,wu2022autocts,han2024bigst} and time-series Transformers~\cite{zhou2021informer,liu2024itransformer,liu2024koopa} have set strong baselines in adjacent domains such as traffic and air quality~\cite{liang2023airformer}, but the former infer connectivity from data with no explicit transport operator, while the latter model temporal dynamics with no spatial structure at all. Even physics-guided neural models that do embed an advective-diffusive prior on a station graph, such as recent air-quality networks~\cite{hettige2024airphynet}, assume dense and roughly uniform monitoring coverage and have not been adapted to a nationwide, density-imbalanced radiation setting with sparse co-located meteorology. Bridging this gap calls for a model that pairs data-driven flexibility with a physical inductive bias for atmospheric transport, learned directly from sparse station observations.

\textbf{A data-driven characterization of the problem.}
Progress has been limited by the absence of a large-scale, publicly available, analysis-ready benchmark. We therefore construct \emph{Japan-1D} and \emph{Japan-4H}, distilled from over four years (March~2021--May~2025) of 10-minute readings from Japan's Nuclear Regulation Authority, covering \textbf{3{,}627} monitoring stations after stringent quality control and aligned with \textbf{228} co-located NOAA-ISD meteorological stations (wind speed, wind direction, air temperature, dew point). The construction pipeline, from quality control to dual-resolution aggregation, is itself a deliverable, and a quantitative analysis of the resulting corpus (\S\ref{sec:data_description_and_analysis}) exposes three structural challenges that jointly defeat off-the-shelf STGNNs and Transformers.

\begin{figure}[t]
  \centering
  \includegraphics[width=1\linewidth]{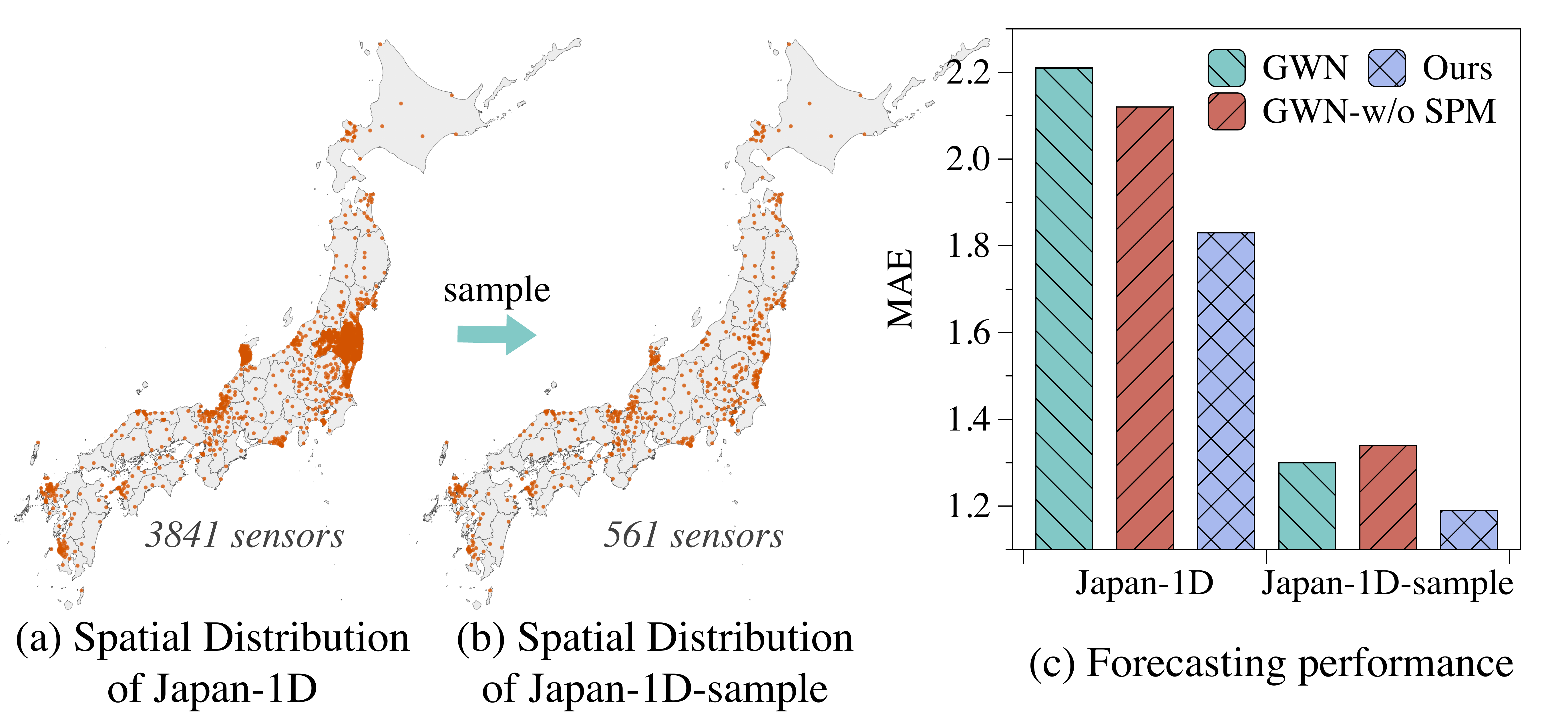}
  \caption{Empirical study of the imbalanced spatial distribution problem. (1)~Full station distribution of Japan-1D. (2)~Evenly-sampled station distribution from Japan-1D. (3)~Forecasting performance of GWN~\cite{wu2019graph} on Japan datasets, where GWN-w/o SPM denotes the variant with the spatial propagation module removed. Removing spatial propagation \emph{improves} accuracy on the full (imbalanced) dataset, indicating that the standard graph operator is net-harmful under extreme station-density imbalance.}
  \label{fig:introduction}
\end{figure}

\textbf{Challenge~1: Non-stationary temporal patterns.} Radiation series are shaped by irregular human interventions, plant-operation events, and the slow decay of long-lived isotopes, producing pronounced distribution shift. On Japan-4H, joint ADF and KPSS tests flag about \textbf{87\%} of stations as non-stationary (\S\ref{sec:temporal_analysis}), and the mean radiation drops by roughly \textbf{3\%} per year, broadly tracking Cs-137 decay and environmental loss overlaid on a strong annual cycle. Temporal modules that assume local stationarity, including many CNN- and RNN-based forecasters~\cite{qiu2024tfb}, can degrade under such shift, and even recent normalization-based Transformers~\cite{kim2021reversible,liu2022non} were not designed for series with decay-driven multi-year trends. A capable model must therefore absorb this distribution shift while staying reactive to sudden events.

\textbf{Challenge~2: Severely imbalanced spatial distribution.} Station density in Japan is dictated by perceived risk rather than geographic coverage. About \textbf{78\%} of the \textbf{3{,}627} stations lie within Fukushima Prefecture and its three immediate neighbors, under \textbf{6\%} of the land surface, so nearest-neighbor distances are highly skewed (\S\ref{sec:spatial_analysis}). Standard message passing over such a graph over-smooths dense clusters and starves sparse regions. As shown in Figure~\ref{fig:introduction}, removing the spatial-propagation module of Graph WaveNet (GWN)~\cite{wu2019graph} actually \emph{improves} accuracy on the full, imbalanced Japan-1D while \emph{hurting} it on an evenly-sampled subset, direct evidence that a one-size-fits-all graph operator is net-harmful under this imbalance. The model must also let spatial information flow by station density rather than through a uniform graph operator.

\textbf{Challenge~3: Physics-grounded heterogeneous context modeling.} Radiation co-evolves with meteorological covariates that are not interchangeable features but quantities governing atmospheric transport. On Japan-4H, station-averaged wind speed and radiation residuals show a weak but consistently located cross-correlation peak (\textbf{16 hours} lag, $|\rho_{\max}|{=}0.16$), pointing to a multi-hour advection timescale rather than instantaneous mixing (\S\ref{sec:meteo_coupling}). The diffusion signature is also wind-dependent. Fitting an isotropic, wind-agnostic spatial Laplacian to radiation increments, we find the diffusion strength declines monotonically with wind speed ($\rho_s{=}{-}1.00$), so this wind-blind operator captures a steadily smaller share of the increments as winds turn the dispersion anisotropic. Recent physics-informed networks~\cite{raissi2019physics,raissi2020hidden,beucler2021enforcing} and physics-aware weather models~\cite{lam2023graphcast,pathak2022fourcastnet} show that injecting such priors is feasible, yet no nationwide radiation forecaster does so. Finally, the model must fuse meteorological context through a physics-grounded inductive bias for wind-modulated transport.

Motivated by these data-driven observations, we introduce \textbf{NRFormer+}, a spatio-temporal Transformer for nationwide nuclear radiation forecasting that substantially extends our conference model NRFormer~\cite{lyu2025nrformer} and integrates three core modules into a single end-to-end pipeline. A \emph{non-stationary temporal attention} module pairs instance-level reversible normalization with point-wise self-attention to remove distribution shift while remaining reactive to sudden events. A \emph{density-adaptive spatial attention} module combines a macro-scale view with a proximity-constrained view, so that information flows differently in dense and sparse regions of the monitoring network. At the center of the model, the new \emph{Physics-Guided Atmospheric Diffusion Module} estimates a meteorology-conditioned diffusion coefficient at every station, approximates how radiation flows from regions of high concentration to neighboring regions of lower concentration over the irregular monitoring graph, and injects this physical signal back into the Transformer pipeline so that learned representations remain consistent with atmospheric transport. These three modules are supported by an enhanced meteorological encoder that routes wind (advection) and thermal (stability and deposition) variables along separate pathways, a day-of-year seasonal embedding that captures annual cycles and slow background decay, a deeper radiation-location cross-feature encoder, and a three-way fusion layer that combines radiation, temporal, and spatial embeddings before forecasting. 
This paper makes the following contributions:
\begin{itemize}
    \item We introduce a \emph{Physics-Guided Atmospheric Diffusion Module} that estimates a meteorology-conditioned diffusion coefficient per station and injects a spatial-Laplacian transport signal into the Transformer as an architectural inductive bias. To our knowledge this is the first learned wind-modulated transport prior for nationwide radiation forecasting, and ablations rank it among the largest contributors to accuracy.
    \item We build \textbf{NRFormer+}, a spatio-temporal Transformer that couples this diffusion prior with non-stationary temporal attention and density-adaptive spatial attention for severely imbalanced sensor networks. It casts nationwide radiation forecasting as graph-structured prediction under a learned transport prior.
    \item We construct and release \textbf{Japan-4H} and \textbf{Japan-1D}, two analysis-ready benchmarks over four years (March~2021--May~2025) with \textbf{3{,}627} radiation stations after stringent quality control and \textbf{228} co-located meteorological stations, to our knowledge the largest publicly available nationwide radiation-forecasting corpora, released with the full construction pipeline.
    \item A quantitative data-engineering analysis establishes the temporal non-stationarity, severe density imbalance, and wind-modulated transport that motivate each module of NRFormer+. Against \textbf{13} strong baselines, NRFormer+ improves consistently across both datasets and all horizons at comparable latency, reducing sudden-change MAE by up to \textbf{19.1\%}.
\end{itemize}

\section{Preliminaries}
\label{sec:Preliminaries}
In this section, we introduce some important definitions and formally define the nuclear radiation forecasting problem.

\begin{myDef}
\textbf{Radiation level}. Radiation level is a number used to quantify the concentrations of radioactive materials in the environment. A higher radiation level indicates that people will experience increasingly detrimental health effects. In practice, the radiation level is computed by using the weighted sum of several radioactive substance measurements, including alpha particles, beta particles, neutron particles, and gamma rays. 
\end{myDef}

\begin{myDef}
\textbf{Radiation monitoring network}. The radiation monitoring network consists of a group of monitoring stations, denoted as $\mathcal{G} = (\mathbf{V}, \mathbf{E})$, where $\mathbf{V}$ is a set of stations and $N=|\mathbf{V}|$ is the number of stations, $\mathbf{E}$ denotes a set of edges representing the relationships among stations. Here we use $\mathbf{A} \in \mathbb{R}^{N \times N}$ to denote the adjacency matrix of the network.
\end{myDef}
To build $\mathcal{G}$, we compute the pairwise distances between stations and derive the adjacency matrix using a pre-defined distance threshold. Let $\mathbf{X} \in \mathbb{R}^{T \times N}$ be the observed radiation level from all stations, where $T$ is the number of time steps. We use $\mathbf{C} \in \mathbb{R}^{T \times N \times C}$ to denote the contextual features associated with each station, \eg meteorological and location information, where $C$ is the feature dimension. Let $\mathcal{H}^{t} = (\mathcal{G}, \mathbf{X}^{t}, \mathbf{C}^{t})$ indicates all the observed values at time step $t$, we define the problem as follows.

\textbf{Nuclear radiation forecasting.}
Given the radiation monitoring network $\mathcal{G}$, historical radiation levels $\mathbf{X}$, contextual features $\mathbf{C}$, the goal is to predict the radiation level for all the monitoring stations over the next $K$ time steps:
\begin{equation}
(\mathbf{\hat{Y}}^{t+1}, \mathbf{\hat{Y}}^{t+2}, \cdots ,\mathbf{\hat{Y}}^{t+K})  \leftarrow \mathcal{F}_{\theta}(\mathcal{H}^{t-P+1}, \mathcal{H}^{t-P+2}, \cdots \mathcal{H}^{t}),
\end{equation}
where $\mathbf{\hat{Y}}^{t+K}$ denotes the predicted value at time step $t+K$, $\mathcal{F}_{\theta}$ is the forecasting model, $P$ and $K$ are the number of historical and future time steps, respectively.

\section{Data Engineering and Analysis}
\label{sec:data_description_and_analysis}

A central premise of this work is that the architectural choices in NRFormer+ are not generic but are dictated by the empirical structure of nationwide radiation data. In this section, we therefore go beyond a descriptive overview and present a quantitative analysis pipeline that (i)~characterizes our newly constructed Japan-1D and Japan-4H datasets and (ii)~establishes, through hypothesis testing and physically motivated statistics, the precise data properties that each component of NRFormer+ is designed to address. 
% The full pipeline (raw observations $\rightarrow$ quality control $\rightarrow$ spatio-temporal alignment $\rightarrow$ dual-resolution aggregation $\rightarrow$ statistical analyses) is detailed in the subsections that follow.

\begin{figure*}[t]
  \centering
  \includegraphics[width=1\linewidth]{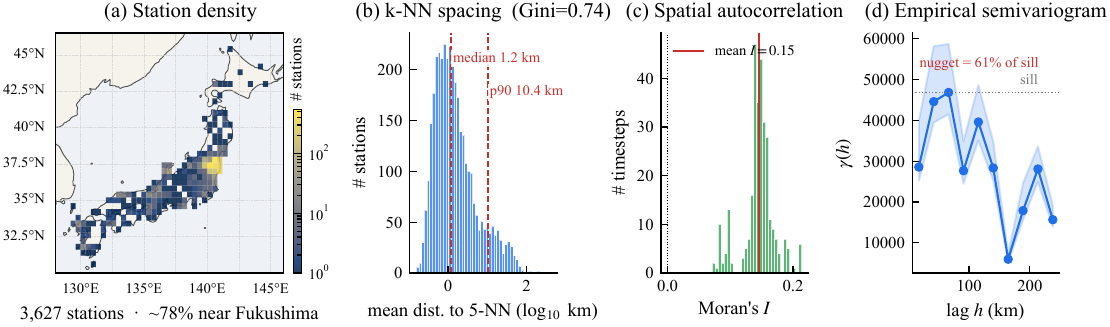}
  \vspace{-0.5cm}
  \caption{Spatial structure of Japan-1D/4H. (a) Hex-binned station density across Japan. (b) Distribution of mean distance to the five nearest neighbors; the Gini coefficient quantifies density imbalance. (c) Time-averaged global Moran's $I$=0.15 on radiation residuals across the test period. (d) Empirical semivariogram, with the nugget at about 61\% of the sill, indicating that most variance is micro-scale rather than spatially structured.}
  \label{fig:spatial_analysis}
\end{figure*}

\subsection{Data Acquisition and Preprocessing}
\label{sec:data_acquisition}

\noindent\textbf{Nuclear radiation data.}
We collected gamma-radiation dose-rate measurements from Japan's Nuclear Regulation Authority (NRA)\footnote{\url{https://www.erms.nsr.go.jp/nra-ramis-webg/}}, comprising over 4{,}000 candidate monitoring stations sampled at a 10-minute cadence from March 17, 2021 to May 31, 2025. Raw measurements were subjected to a four-stage quality-control pipeline, (1)~station-level pruning that removes any station with more than 30 days of cumulative missing data, retaining \textbf{3{,}627} reliable stations out of the 3{,}841 used in the~\cite{lyu2025nrformer}; (2)~outlier suppression via a Hampel filter ($k{=}3$, window $w{=}24$ samples) that flags physically implausible spikes inconsistent with adjacent stations; (3)~gap imputation through temporally local linear interpolation for short gaps ($\leq$2\,h) and station-conditional seasonal decomposition for longer gaps; and (4)~dual-resolution aggregation, producing the 4-hour resolution \emph{Japan-4H} (9{,}222 timesteps, supporting short-term operational forecasting) and the daily resolution \emph{Japan-1D} (1{,}537 timesteps, supporting long-term strategic planning).

\noindent\textbf{Meteorological data.}
We integrated NOAA's Integrated Surface Dataset (ISD)\footnote{\url{https://www.ncei.noaa.gov/metadata/geoportal/rest/metadata/item/gov.noaa.ncdc:C00532/html}}, retaining \textbf{228} weather stations (after removing those with insufficient coverage) that were geographically matched to the radiation network. We extracted four physically motivated variables, wind speed and wind direction (governing advective transport), air temperature (modulating atmospheric stability), and dew point (a proxy for moisture content controlling wet deposition). All meteorological series were resampled onto the radiation grid through Akima spline interpolation, which preserves local monotonicity better than cubic interpolation for meteorological variables.

\subsection{Dataset Statistics}
\label{sec:dataset_compare}

Table~\ref{tab:dataset_description} summarizes the Japan-1D/4H statistics. The datasets cover \textbf{3{,}627} monitoring stations paired with \textbf{228} co-located meteorological stations over four years, with four physically motivated meteorological covariates (wind speed, wind direction, air temperature, dew point) and a strongly skewed station-density distribution. To our knowledge, Japan-1D/4H constitute the largest publicly released nationwide nuclear-radiation forecasting corpus.

\begin{table}[t] \small
  \caption{Statistics of the radiation and meteorological datasets used in NRFormer+.}
  \label{tab:dataset_description}
  \begin{tabular}{cl|cc}
    \toprule
    \multicolumn{2}{c|}{Data Description} & Japan-4H & Japan-1D \\
    \midrule
    \multirow{4}{*}{\text { Nuclear radiation data }}
    & \# of stations & 3{,}627 & 3{,}627 \\
    & \# of timesteps & 9{,}222 & 1{,}537 \\
    & Interval & 4 hour & 1 day \\
    & Time span & \multicolumn{2}{c}{3/17/2021 -- 5/31/2025} \\
    \midrule
    \multirow{4}{*}{\text { Meteorological data }}
    & \# of stations & \multicolumn{2}{c}{228} \\
    & \# of timesteps &  \multicolumn{2}{c}{36{,}888} \\
    & Interval &  \multicolumn{2}{c}{1 hour} \\
    & Time span &  \multicolumn{2}{c}{3/17/2021 -- 5/31/2025} \\
    \bottomrule
  \end{tabular}
\end{table}

\subsection{Spatial Analysis: Imbalanced Density and Heterogeneous Correlations}
\label{sec:spatial_analysis}

\noindent\textbf{Finding S1 (Imbalanced station density).}
Approximately $78\%$ of the 3{,}627 stations are concentrated within Fukushima Prefecture and its three immediate neighbors, which collectively occupy less than $6\%$ of Japan's land area. The empirical $k$-NN distance distribution ($k{=}5$) is heavily right-skewed. The Gini coefficient of mean inter-station distances is $0.74$, and the median distance to the five nearest neighbors is only $1.2$\,km, while the 90th percentile reaches $10.4$\,km, indicating an order-of-magnitude spread (Figure~\ref{fig:spatial_analysis}(b)). When measured by the densest decile of the spatial domain, $83.6\%$ of all stations fall within only $10\%$ of the area, an imbalance ratio that far exceeds that of standard sensor-network benchmarks.

\noindent\textbf{Finding S2 (Heterogeneous spatial correlation).}
Standard graph-based STGNNs implicitly assume Tobler's first law, \ie that geographic proximity implies similarity. We test this assumption by computing the global Moran's $I$ statistic on radiation residuals using a Gaussian-kernel weight matrix, obtaining $I{=}0.15$, indicating positive but modest spatial autocorrelation. The empirical semivariogram (Figure~\ref{fig:spatial_analysis}(d)) exhibits a pronounced \emph{nugget effect} of $\approx\!61\%$ of the total sill, signalling substantial micro-scale variability that cannot be explained by distance alone, and an effective correlation range of only $\approx\!66$\,km.

% \noindent\textbf{Implication.}
Findings S1 and S2 jointly invalidate uniform-neighborhood graph propagation. Dense-region nodes are saturated by redundant signals, sparse-region nodes are informationally starved, and even nearby pairs may be weakly correlated. This directly motivates the \emph{Density-Adaptive Spatial Attention} (Section~\ref{sec:spatial_attention}), which routes attention based on learned radiation similarity rather than geographic distance.

\subsection{Temporal Analysis: Non-stationarity, Multi-scale Periodicity, and Heavy-tailed Events}
\label{sec:temporal_analysis}

\begin{figure}[t]
  \centering
  \includegraphics[width=1\linewidth]{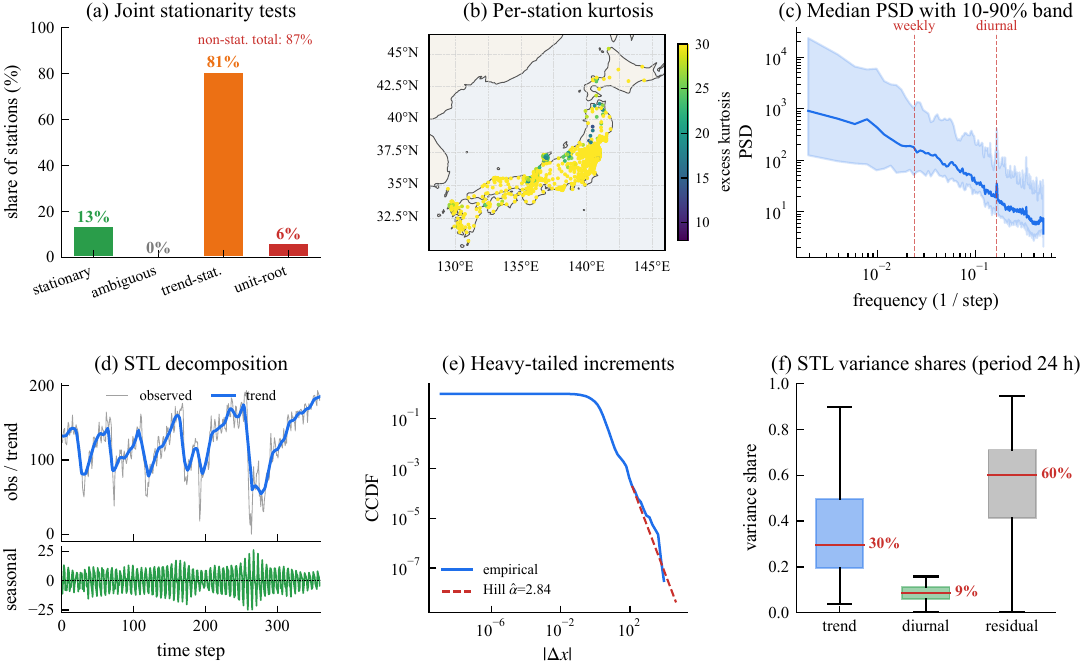}
  \caption{Temporal characterization of nationwide radiation series. (a) Joint ADF and KPSS stationarity outcomes across stations, with about 87\% non-stationary. (b) Per-station excess kurtosis projected onto Japan's geography. (c) Median Welch power spectral density with 10--90\% inter-station band; vertical lines mark diurnal, weekly and annual frequencies. (d) Representative STL decomposition. (e) Pooled CCDF of first-difference magnitudes with Hill-estimator tail slope $\hat{\alpha}$. (f) Per-station variance attribution to trend, seasonal, and residual.}
  \label{fig:temporal_analysis}
\end{figure}

\noindent\textbf{Finding T1 (Pervasive non-stationarity).}
We applied the Augmented Dickey--Fuller (ADF)~\cite{dickey1979distribution} and Kwiatkowski--Phillips--Schmidt--Shin (KPSS)~\cite{kpss1992testing} tests to every station-level series and classify each station by the joint outcome (Figure~\ref{fig:temporal_analysis}(a)). At significance level $\alpha{=}0.05$, only $\sim\!13\%$ of stations pass both tests as stationary, $\sim\!81\%$ reject KPSS while passing ADF (trend-stationary with shifts), and an additional $\sim\!6\%$ fail both (unit root with shifts), totalling $\sim\!87\%$ non-stationary. Per-station excess kurtosis (Fig.~\ref{fig:temporal_analysis}b) exceeds 3 for more than $80\%$ of stations and reaches values above 20 near Fukushima Daiichi; variance decomposition (panel f) attributes the bulk of residual variance to high-frequency components, supporting a non-Gaussian, event-driven temporal regime. A near-linear downward trend of approximately $2.9\%$ per annum, captured well across stations ($R^2{=}0.56$), is consistent with Cs-137 decay augmented by environmental loss processes such as weathering and runoff.

\noindent\textbf{Finding T2 (Multi-scale periodicity).}
Welch power-spectral analysis (Figure~\ref{fig:temporal_analysis}(c)) and STL decomposition (Figure~\ref{fig:temporal_analysis}(d)) jointly reveal a hierarchy of cyclic components, a \emph{circadian} cycle ($T{=}24$\,h, captured only by Japan-4H) accounting for $\approx\!4.5\%$ of station-level variance, a secondary peak at $\sim$7 days plausibly aligned with synoptic weather oscillations, and a dominant \emph{annual} cycle accounting for $\approx\!35\%$ of variance, with January exhibiting the lowest mean radiation.

\noindent\textbf{Finding T3 (Heavy-tailed sudden-change events).}
The complementary cumulative distribution of first-order increments $|\Delta C_t|$ deviates substantially from a Gaussian baseline (Figure~\ref{fig:temporal_analysis}(e)), a Hill-estimator fit on the upper $5\%$ tail yields an exponent of $\hat{\alpha}{=}2.84$, and over $99\%$ of stations exhibit at least one $>$5$\sigma$ jump in the analysis window. This heavy-tailed behavior is consistent with episodic, weather-triggered transport events overlaid on the smooth decay trend.

% \noindent\textbf{Implication.}
T1 motivates the \emph{Non-stationary Temporal Attention} with instance-level normalization (Section~\ref{sec:temporal_attention}), which absorbs distribution shifts before attention. T2 justifies the day-of-year seasonal embedding in the Context-Aware Propagation Prompting module (Section~\ref{sec:temporal_encoder}). T3 indicates that the model must remain reactive to abrupt transport events, a property that purely smooth recurrent models lack and that the point-wise temporal attention together with the physics-guided Laplacian term in Section~\ref{sec:diffusion_module} explicitly preserves.

\subsection{Meteorology--Radiation Coupling: Empirical Evidence for Atmospheric Transport}
\label{sec:meteo_coupling}

This subsection presents the empirical evidence that anchors the new \emph{Physics-Guided Atmospheric Diffusion Module}. The advection--diffusion equation~\eqref{eq:adv-diff-intro} (Section~\ref{sec:diffusion_module}) predicts two falsifiable signatures, (i) radiation responses should lag wind events, and (ii) the effective diffusion observed under an isotropic graph operator should be modulated by wind, since strong winds redirect transport from omnidirectional mixing to anisotropic advection.

\begin{figure}[t]
  \centering
  \includegraphics[width=1\linewidth]{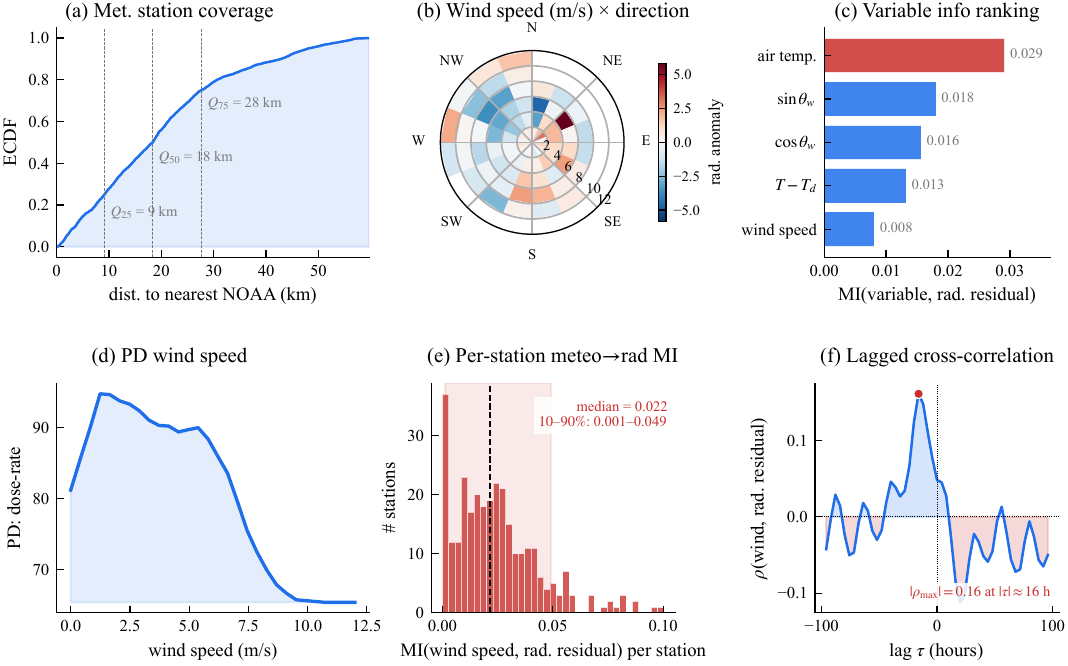}
  \caption{Meteorology--radiation coupling. (a) ECDF of each radiation station's distance to its nearest NOAA station. (b) Wind-speed $\times$ wind-direction conditional radiation anomaly (per-station de-meaned). (c) Mutual-information ranking of meteorological variables against the per-station radiation residual. (d) Partial-dependence of radiation on wind speed from a gradient-boosted surrogate. (e) Distribution of per-station mutual information between wind speed and the radiation residual. (f) Lagged cross-correlation of station-averaged wind speed and radiation residual, peaking at a multi-hour lag.}
  \label{fig:meteo_coupling}
\end{figure}

\noindent\textbf{Finding M1 (Lagged response to wind events).}
The two-sided cross-correlation between station-averaged wind speed and radiation residuals attains $|\rho_{\max}|{=}0.16$ at a lag whose magnitude is approximately $16$\,h on Japan-4H. The non-zero, multi-hour lag is incompatible with instantaneous mixing models and instead supports a transport-and-diffusion picture in which wind events drive distributed dispersion over a multi-hour timescale (Figure~\ref{fig:meteo_coupling}(f)). Beyond the existence of the lag, two complementary signatures further constrain the encoder design. The per-station mutual information between wind speed and the radiation residual (Figure~\ref{fig:meteo_coupling}(e)) is widely dispersed across the $3{,}627$ stations, indicating that the strength of the meteorology$\to$radiation coupling is itself station-specific, and the variable-level information ranking (Figure~\ref{fig:meteo_coupling}(c)) shows that air temperature and the wind direction components ($\sin\theta_w$, $\cos\theta_w$) carry as much signal as wind speed, justifying a multi-channel rather than wind-speed-only meteorological pathway. The wind-direction-conditioned radiation anomaly is moreover strongly asymmetric (Figure~\ref{fig:meteo_coupling}(b)), a directional effect confirmed by a gradient-boosted partial-dependence analysis (Figure~\ref{fig:meteo_coupling}(d)), supporting explicit treatment of advection.

\begin{figure*}[t]
  \centering
  \includegraphics[width=1\linewidth]{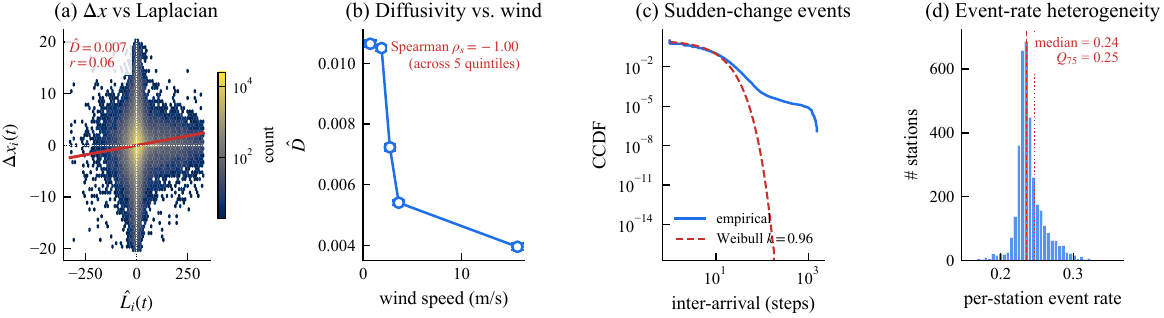}
  \vspace{-0.4cm}
  \caption{Empirical validation of advection--diffusion priors. (a) Hex-binned scatter of the first-difference $\Delta x_i(t)$ against the discrete Laplacian $\hat{L}_i(t)$ on the $k{=}10$ graph, with a small positive fit slope ($\hat{D}>0$, $r{=}0.06$). (b) $\hat{D}$ stratified by wind-speed quintile, decreasing monotonically with wind speed ($\rho_s{=}{-}1.00$). (c) CCDF of inter-arrival times of $>3\cdot$MAD sudden-change events with a Weibull fit. (d) Per-station sudden-change event rates (median 0.24, $Q_{75}$ 0.25).}
  \label{fig:physics_validation}
  \vspace{-0.4cm}
\end{figure*}

\noindent\textbf{Finding M2 (Wind-modulated effective diffusion).}
We estimated an empirical effective diffusion coefficient $\hat{D}$ by regressing first-order increments $\Delta x_i(t)$ onto the discrete spatial Laplacian $\hat{L}_i(t)$ on the $k$=10 NN graph (Figure~\ref{fig:physics_validation}(a)) and stratified the regression slope by wind-speed quintile. The fitted slope $\hat{D}$ decreases monotonically across all five wind-speed quintiles (Figure~\ref{fig:physics_validation}(b), Spearman $\rho_s{=}{-}1.00$). The intuitive interpretation is that the isotropic graph Laplacian explains a smaller share of $\Delta x$ when wind drives anisotropic transport. Simple isotropic diffusion is therefore an inadequate operator at high wind, motivating a learned, wind-modulated diffusion in which the operator and coefficient are jointly conditioned on meteorological state. The resulting transport manifests as episodic sudden-change events, whose inter-arrival times follow a near-memoryless Weibull law (Figure~\ref{fig:physics_validation}(c)) and whose per-station rates vary widely across the network (Figure~\ref{fig:physics_validation}(d)).

% \noindent\textbf{Implication.}
M1 and M2 jointly establish that radiation dynamics in Japan-1D/4H are coupled to meteorology in a manner that is both lagged and anisotropic, and that an isotropic, wind-agnostic diffusion operator under-explains the observed transport. This is precisely the inductive bias encoded in the \emph{Physics-Guided Atmospheric Diffusion Module} (Section~\ref{sec:diffusion_module}), which estimates a station-wise $D_i$ from the local meteorological context, and in the \emph{Enhanced Meteorological Encoder} (Section~\ref{sec:meteo_encoder}), which separates wind (advection) and thermal (stability) pathways.

\subsection{Implications for Model Design}
\label{sec:implications}

In summary, the data engineering and analysis pipeline described above transforms nationwide radiation and meteorological observations into a rigorously characterized forecasting corpus, and the resulting empirical signatures (non-stationarity, multi-scale periodicity, heavy-tailed transients, imbalanced and heterogeneous spatial structure, and meteorology-modulated diffusion) collectively define the design space within which NRFormer+ operates. We next describe how each of these findings is translated into a concrete architectural mechanism.

\begin{figure*}[t]
  \centering
  \includegraphics[width=1\linewidth]{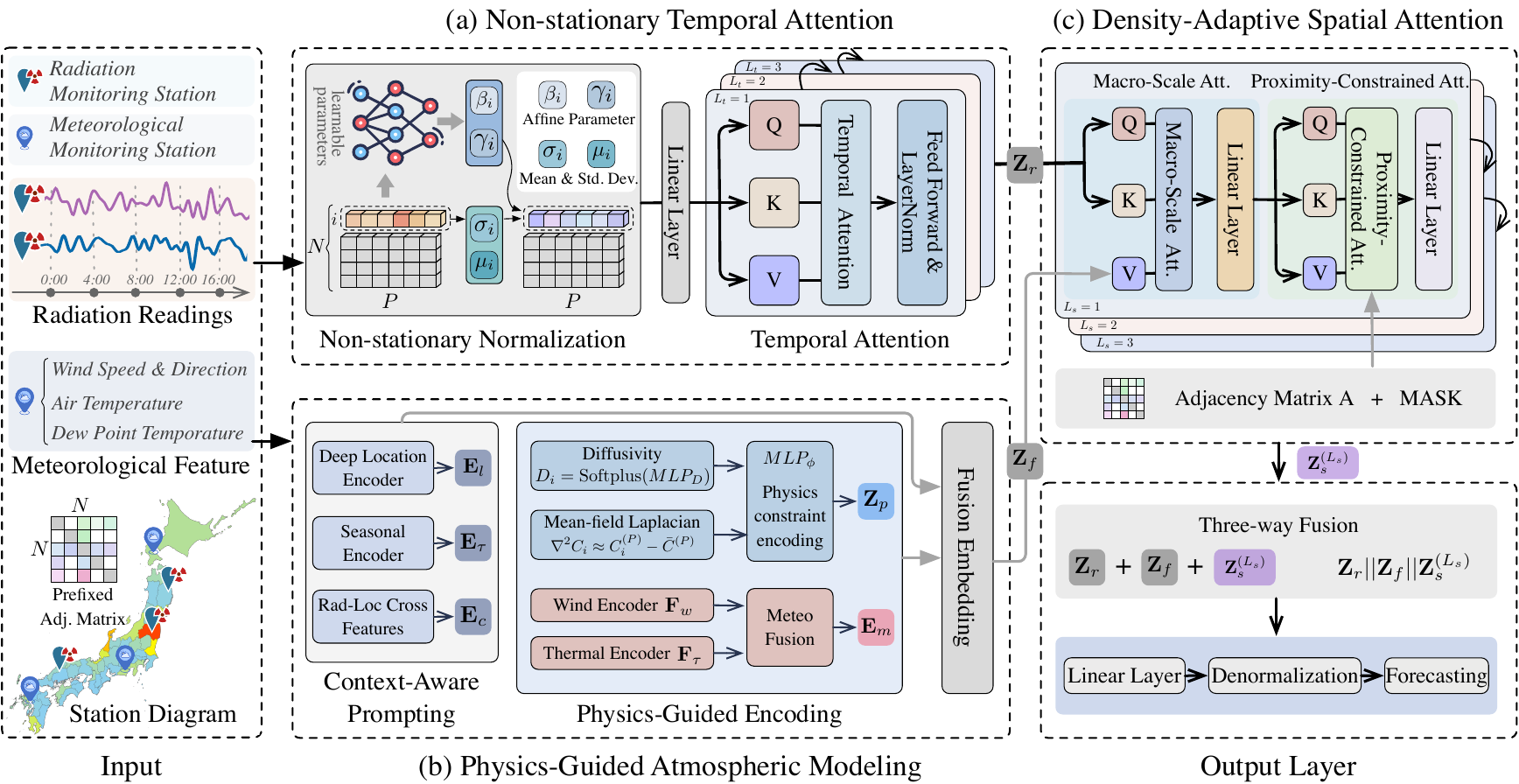}
  % \vspace{-0.5cm}
  \caption{The framework overview of NRFormer+.}
  \label{fig:framework}
  % \vspace{-0.2cm}
\end{figure*}

\section{Methodology}

\textbf{Overview.}
Building on NRFormer~\cite{lyu2025nrformer}, which relied on purely data-driven temporal and spatial attention, NRFormer+ adds explicit mechanisms to encode the physics of atmospheric radiation transport. Its overall framework, the Physics-Guided Radiation Transformer, is illustrated in Figure~\ref{fig:framework}.

% Compared with NRFormer, NRFormer+ introduces the following key innovations:
% (1) \textit{Physics-Guided Atmospheric Modeling:} a novel atmospheric diffusion module that explicitly encodes the governing equation of radiation transport (\ie the diffusion equation $\partial C / \partial t = D \nabla^2 C$), along with an enhanced meteorological encoder that separately models wind dynamics and thermal conditions through physically-motivated architectural designs.
% (2) \textit{Context-Aware Propagation Prompting:} a substantially expanded prompting mechanism that incorporates a deep location encoder, a seasonal temporal encoder based on day-of-year embeddings, and radiation-location cross features that capture the joint distribution of radiation patterns and geographical positions.
% (3) \textit{Physics-Enriched Spatial Attention:} a redesigned spatial attention mechanism where radiation features directly guide the query-key routing, enabling the model to discover spatial dependencies grounded in actual radiation patterns rather than solely temporal embeddings.
% (4) \textit{Three-Way Fusion Output:} an enhanced output layer that preserves a direct radiation signal path alongside the temporal and spatial pathways, ensuring robust gradient flow and preventing the loss of raw radiation information through successive transformations.

% NRFormer+ retains the non-stationary temporal attention and imbalance-aware spatial attention from NRFormer while substantially enriching the model with physics-guided inductive biases. We detail each component below.

\subsection{Non-stationary Temporal Attention}
\label{sec:temporal_attention}
Radiation time series exhibit a highly non-stationary nature due to unpredictable events such as irregular human interventions and nuclear power plant operational changes. To effectively extract stable knowledge from radiation sequence inputs, we employ a non-stationary temporal attention mechanism that consists of two components, non-stationary normalization and point-wise temporal attention.

\subsubsection{Non-stationary Normalization}
Due to the non-stationarity of radiation time series, the underlying distributions of different input sequences are diverse, which significantly degrades the forecasting performance. Motivated by recent works~\cite{kim2021reversible,liu2022non}, we apply instance-wise normalization to eliminate the non-stationary information in each input sequence.
Formally, given historical observations $\mathbf{x}_{i} \in \mathbb{R}^{P}$ at the $i$-th station, where $P$ is the input time window, we first compute the instance-level statistics:
\begin{equation}
\begin{aligned}
\mathbb{E}[\mathbf{x}_{i}] = \frac{1}{P}\sum_{j=1}^{P}\mathbf{x}^{j}_{i},\quad Var[\mathbf{x}_{i}] = \frac{1}{P}\sum_{j=1}^{P}(\mathbf{x}^j_{i}-\mathbb{E}[\mathbf{x}_{i}])^{2}.
\end{aligned}
\end{equation}
We then normalize the original input sequence with learnable affine parameters:
\begin{equation}
\begin{aligned}
\hat{\mathbf{x}}_{i} = \gamma_i \frac{\mathbf{x}_{i}-\mathbb{E}[\mathbf{x}_{i}]} {\sqrt{Var[\mathbf{x}_{i}] + \epsilon}} + \beta_i,
\end{aligned}
\end{equation}
where $\gamma_i$ and $\beta_i$ are learnable parameters corresponding to the $i$-th station, and $\epsilon$ is a small constant for numerical stability. The normalized sequences possess more stable statistical properties, substantially reducing the difficulty of capturing temporal dynamics.
Subsequently, we embed $\hat{\mathbf{x}}_{i}$ through a $1\times 1$ convolutional layer:
$\mathbf{z}_{t}^{i} = \text{Conv}_{1\times 1}(\hat{\mathbf{x}}_{i})$,
where $\mathbf{z}_{t}^{i} \in \mathbb{R}^{P \times D}$ represents the point-wise time series embedding of station $i$, and $D$ is the hidden dimension.

\subsubsection{Point-wise Temporal Attention}
To capture both unstable and long-range temporal dependencies inherent in nuclear radiation time series, we employ a point-wise temporal self-attention mechanism. Given the embedding $\mathbf{z}_{t}^{i}$, we derive query, key, and value matrices $\mathbf{Q}_t = \mathbf{z}_{t}^{i} \mathbf{W}_t^Q$, $\mathbf{K}_t = \mathbf{z}_{t}^{i} \mathbf{W}_t^K$, and $\mathbf{V}_t = \mathbf{z}_{t}^{i} \mathbf{W}_t^V$, where $\mathbf{W}_t^Q, \mathbf{W}_t^K, \mathbf{W}_t^V \in \mathbb{R}^{D\times D}$ are learnable parameters shared across stations. The temporal attention score is computed as
$\mathbf{A}_t = \text{Softmax}\!\left(\frac{\mathbf{Q}_t {\mathbf{K}_t}^{\top}}{\sqrt{D}}\right)$,
where $\mathbf{A}_t \in \mathbb{R}^{P\times P}$. The updated representation is obtained via $\mathbf{z}_{t}^{i,(l)} = \mathbf{A}_t \mathbf{V}_t$.
We employ multi-head attention with residual connections and layer normalization~\cite{vaswani2017attention}, stacking $L_t$ such layers. The final temporal output is obtained by flattening and projecting:
$\mathbf{z}_{t}^{i} = \text{Flatten}(\mathbf{z}_{t}^{i,(L_t)}) \mathbf{W}^{r}$,
where $\mathbf{W}^{r} \in \mathbb{R}^{(P \cdot D)\times D}$. We denote $\mathbf{Z}_r \in \mathbb{R}^{N\times D}$ as the radiation feature matrix of all stations, where the $i$-th row is $\mathbf{z}_{t}^{i}$.

\subsection{Physics-Guided Atmospheric Modeling}
\label{sec:physics_modeling}
A fundamental limitation of purely data-driven approaches for radiation forecasting is their inability to encode the physical laws governing atmospheric transport. Nuclear radiation disperses through the atmosphere primarily via diffusion and advection processes, which are well-characterized by established partial differential equations. In NRFormer+, we introduce two physics-guided components that inject domain knowledge into the learning process, an atmospheric diffusion module grounded in the diffusion equation, and an enhanced meteorological encoder with physically-motivated architectural designs.

\subsubsection{Atmospheric Diffusion Module}
\label{sec:diffusion_module}
The atmospheric transport of radioactive materials is fundamentally governed by the advection--diffusion equation~\cite{stockie2011mathematics}:
\begin{equation}
\frac{\partial C}{\partial t} \;=\; -\mathbf{u}\cdot\nabla C \;+\; \nabla\!\cdot(D\,\nabla C) \;+\; S,
\label{eq:adv-diff-intro}
\end{equation}
where $C$ is the radiation concentration, $\mathbf{u}$ is the wind field, $D$ is an eddy-diffusion coefficient that depends on atmospheric stability, and $S$ aggregates sources and sinks. In NRFormer+, the advective term $-\mathbf{u}\cdot\nabla C$ is absorbed by the density-adaptive spatial attention mechanism (Section~\ref{sec:spatial_attention}), which routes information by learned radiation similarity over the irregular monitoring graph. Under this decomposition, the diffusive component of Eq.~\eqref{eq:adv-diff-intro} reduces to the diffusion equation $\frac{\partial C}{\partial t} = D \nabla^2 C$, where $\nabla^2 C$ is the spatial Laplacian of the concentration field. Rather than attempting to numerically solve this PDE, we design a neural module that encodes its key physical quantities as inductive biases.

\textbf{Diffusion coefficient estimation.}
The diffusion coefficient $D$ is not a universal constant but varies with local atmospheric conditions, including wind turbulence, atmospheric stability, and terrain characteristics. We estimate $D$ for each station using a parameterized neural network that takes meteorological and geographical features as inputs:
\begin{equation}
\label{equ:diffusion_coeff}
D_i = \text{Softplus}\!\left(\text{MLP}_D\!\left([\bar{\mathbf{m}}_i \| \mathbf{l}_i]\right)\right),
\end{equation}
where $\bar{\mathbf{m}}_i \in \mathbb{R}^{C_m}$ is the time-averaged meteorological feature vector at station $i$ (\ie the mean over the $P$ input time steps), $\mathbf{l}_i \in \mathbb{R}^{2}$ is the geographic coordinate (latitude, longitude), and $\|$ denotes concatenation. The $\text{MLP}_D$ consists of three fully-connected layers with ReLU activations ($\mathbb{R}^{C_m+2} \rightarrow \mathbb{R}^{64} \rightarrow \mathbb{R}^{32} \rightarrow \mathbb{R}^{1}$). Critically, the Softplus activation function ensures that $D_i > 0$, which is a physical constraint since diffusion coefficients must be strictly positive.

\textbf{Spatial Laplacian approximation.}
Computing the exact Laplacian $\nabla^2 C$ requires the continuous concentration field, which is observed only at discrete stations. For the diffusion module we therefore use a lightweight surrogate that needs no graph, namely each station's deviation from the global spatial mean, $\nabla^2 C_i \approx C_i^{(P)} - \bar{C}^{(P)}$, where $C_i^{(P)}$ is the radiation level at station $i$ at the most recent time step $P$ and $\bar{C}^{(P)} = \frac{1}{N}\sum_{j=1}^{N} C_j^{(P)}$ is the spatial mean. This is a zeroth-order, mean-field coarsening of the local graph Laplacian~\cite{shuman2013emerging}, a global-anomaly signal rather than a true local curvature; we adopt it because it needs no neighbor graph and stays informative on the heavy-tailed events where a station departs sharply from the network background. The corresponding local diffusive tendency, measured with a $k$-NN graph Laplacian, is the quantity empirically validated in Finding M2 (\S\ref{sec:meteo_coupling}).

\textbf{Physics constraint encoding.}
Having obtained the three key physical quantities from the diffusion equation, namely the concentration $C_i^{(P)}$, the estimated diffusion coefficient $D_i$, and the mean-field Laplacian surrogate $\nabla^2 C_i$, we encode this physics triplet into a latent representation:
\begin{equation}
\label{equ:physics_encoding}
\mathbf{p}_i = \text{MLP}_\phi\!\left(\left[C_i^{(P)},\; D_i,\; \nabla^2 C_i\right]\right),
\end{equation}
where $\text{MLP}_\phi: \mathbb{R}^{3} \rightarrow \mathbb{R}^{D}$ is a two-layer MLP with ReLU activation. The resulting physics embedding $\mathbf{p}_i \in \mathbb{R}^{D}$ encodes the local diffusive tendency at each station. Let $\mathbf{Z}_p \in \mathbb{R}^{N\times D}$ denote the embeddings for all stations, where $i$-th row is $\mathbf{p}_i$.

Encoding the triplet as a soft feature pathway rather than a hard penalty term tells the model which stations show concentration anomalies (via the Laplacian term) and how fast they should dissipate (via $D_i$), while still allowing non-diffusive patterns such as discrete emission events to be learned.

\subsubsection{Enhanced Meteorological Encoder}
\label{sec:meteo_encoder}
In the conference version NRFormer, meteorological features were processed through flat concatenation followed by a single MLP, treating all meteorological variables homogeneously. However, different meteorological variables govern fundamentally different physical processes in radiation transport. Wind drives advective transport, while temperature conditions control atmospheric stability and vertical mixing. NRFormer+ introduces a physically-motivated encoder that processes these variable groups through separate pathways before fusion.

\textbf{Wind dynamics encoder.}
Wind speed and direction jointly determine the advective transport of radioactive particles. We encode wind features using a two-stage architecture:
\begin{equation}
\label{equ:wind_encoder}
\mathbf{F}_w = \text{Conv}^{(2)}_{1\times 3}\!\left(\text{ReLU}\!\left(\text{Conv}^{(1)}_{1\times 1}(\mathbf{M}_w)\right)\right),
\end{equation}
where $\mathbf{M}_w \in \mathbb{R}^{B \times 2 \times N \times P}$ contains the wind speed and direction measurements across all stations and time steps. The first $1\times 1$ convolution projects the two wind channels into $D/2$ feature channels, capturing the interaction between speed and direction (\eg computing effective wind velocity components). The second $1\times 3$ temporal convolution with padding captures the short-term dynamics of wind patterns. The output $\mathbf{F}_w \in \mathbb{R}^{B \times (D/2) \times N \times P}$ encodes the wind dynamics.

\textbf{Thermal condition encoder.}
Air temperature and dew point jointly characterize atmospheric stability, which determines the vertical extent of radiation mixing. We employ an analogous architecture:
\begin{equation}
\label{equ:temp_encoder}
\mathbf{F}_\tau = \text{Conv}^{(2)}_{1\times 3}\!\left(\text{ReLU}\!\left(\text{Conv}^{(1)}_{1\times 1}(\mathbf{M}_\tau)\right)\right),
\end{equation}
where $\mathbf{M}_\tau \in \mathbb{R}^{B \times 2 \times N \times P}$ contains air temperature and dew point measurements. The temperature-dew point differential indicates atmospheric stability, a physically meaningful quantity the convolutional architecture can learn to extract.

\textbf{Fusion.}
The wind and thermal features are concatenated and transformed through an MLP with residual connections to produce the final meteorological embedding:
\begin{equation}
\label{equ:meteo_fusion}
\mathbf{E}_m = \text{MLP}_{f}\!\left(\text{Flatten}\!\left([\mathbf{F}_w \| \mathbf{F}_\tau]\right)\right),
\end{equation}
where $\text{Flatten}(\cdot)$ collapses the temporal and channel dimensions, and $\text{MLP}_{f}$ consists of a $1\times 1$ convolution followed by two residual MLP blocks and a final projection to $\mathbb{R}^{D}$. The resulting $\mathbf{E}_m \in \mathbb{R}^{N\times D}$ provides a comprehensive meteorological representation for each station.

\subsubsection{Context-Aware Propagation Prompting}
\label{sec:propagation_prompting}
The propagation of radiation is influenced by a complex interplay of geographical, temporal, and meteorological factors. While NRFormer incorporated basic location and meteorological prompts via concatenation into the spatial attention's query and key, NRFormer+ substantially expands the prompting mechanism with three additional context encoders that capture richer aspects of the radiation propagation environment.

\paragraph{Deep Location Encoder}
\label{sec:location_encoder}
Different geographic locations exhibit distinct radiation characteristics due to variations in terrain, proximity to nuclear facilities, and local atmospheric circulation patterns. We encode station coordinates through a deep MLP that can learn nonlinear spatial embeddings:
\begin{equation}
\label{equ:location_encoder}
\mathbf{E}_{l} = \text{MLP}_l(\mathbf{L}),\quad \text{MLP}_l: \mathbb{R}^{2} \xrightarrow{\text{ReLU}} \mathbb{R}^{32} \xrightarrow{\text{ReLU}} \mathbb{R}^{64} \xrightarrow{} \mathbb{R}^{D},
\end{equation}
where $\mathbf{L} \in \mathbb{R}^{N\times 2}$ contains the latitude and longitude of each station. Compared to the simpler two-layer encoder in NRFormer, the three-layer architecture with progressive dimension expansion ($2 \rightarrow 32 \rightarrow 64 \rightarrow D$) enables the model to capture more complex spatial patterns, such as the influence of mountain ranges on radiation shielding or coastal effects on atmospheric dispersion. The location embedding $\mathbf{E}_{l} \in \mathbb{R}^{N\times D}$ is shared across all samples in a batch.

\paragraph{Seasonal Temporal Encoder}
\label{sec:temporal_encoder}
As demonstrated in our data analysis (Section~\ref{sec:data_description_and_analysis}), radiation levels exhibit pronounced seasonal variations, with January consistently showing the lowest values across the country. NRFormer encoded temporal context through three separate embeddings for time-of-day, day-of-week, and month-of-year. NRFormer+ replaces these with a unified day-of-year embedding that directly captures the continuous seasonal cycle:
\begin{equation}
\label{equ:temporal_encoder}
\mathbf{E}_\tau = \text{Embed}_\text{doy}(d),\quad \text{Embed}_\text{doy}: \{1, 2, \ldots, 366\} \rightarrow \mathbb{R}^{D},
\end{equation}
where $d$ denotes the day-of-year index extracted from the last time step of the input window. The embedding table $\text{Embed}_\text{doy} \in \mathbb{R}^{366 \times D}$ is initialized with Xavier uniform initialization and learned end-to-end. This design offers two advantages over separate time embeddings, (1) it provides a single, coherent representation of seasonal position rather than forcing the model to reconstruct seasonality from independent components, and (2) it increases representational specificity, since each day-of-year has its own unique embedding. The embedding $\mathbf{E}_\tau \in \mathbb{R}^{N\times D}$ is broadcast across all stations.

\paragraph{Radiation-Location Cross Features}
\label{sec:cross_features}
While the location encoder and radiation features provide complementary perspectives, their interaction may reveal important patterns. For instance, stations at similar latitudes near coastal areas may share radiation dynamics that differ from inland stations at the same latitude. We introduce a cross-feature module that models the joint distribution of radiation and location:
\begin{equation}
\label{equ:cross_features}
\mathbf{R}_\text{loc} = [\mathbf{X}_{i} \| \mathbf{L}^{(P)}_{i,\text{lat}} \| \mathbf{L}^{(P)}_{i,\text{lon}}] \in \mathbb{R}^{N\times 3P},
\end{equation}
where $\mathbf{X}_{i} \in \mathbb{R}^{P}$ is the radiation time series at station $i$, and $\mathbf{L}^{(P)}_{i,\text{lat}}, \mathbf{L}^{(P)}_{i,\text{lon}} \in \mathbb{R}^{P}$ are the latitude and longitude coordinates replicated across the $P$ time steps. This concatenation is processed through a $1\times 1$ convolution followed by three residual MLP blocks:
\begin{equation}
\label{equ:cross_encoder}
\mathbf{E}_c = \text{ResMLP}^{(3)}\!\left(\text{Conv}_{1\times 1}(\mathbf{R}_\text{loc})\right),
\end{equation}
where each residual MLP block consists of two fully-connected layers with ReLU activation and a skip connection. The resulting cross embedding $\mathbf{E}_c \in \mathbb{R}^{N\times D}$ captures how radiation patterns co-vary with geographic position.

\paragraph{Multi-Source Temporal Fusion}
All contextual embeddings are aggregated with the radiation and physics features through a temporal fusion layer:
\begin{equation}
\label{equ:temporal_fusion}
\mathbf{Z}_f = \text{Conv}_{1\times 1}\!\left([\mathbf{Z}_r \| \mathbf{Z}_p \| \mathbf{E}_l \| \mathbf{E}_c \| \mathbf{E}_m \| \mathbf{E}_\tau]\right),
\end{equation}
where the $1\times 1$ convolution projects the concatenated $(6D)$-dimensional representation back to $D$ dimensions. The fused embedding $\mathbf{Z}_f \in \mathbb{R}^{N\times D}$ serves as the value input $\mathbf{V}_s$ to the spatial attention mechanism, thereby providing physics-enriched contextual information for spatial message passing.

\subsection{Density-Adaptive Spatial Attention}
\label{sec:spatial_attention}
As analyzed in Section~\ref{Introduction}, the severely imbalanced spatial distribution of monitoring stations poses a significant challenge. Nodes in dense clusters risk over-smoothing from excessive information aggregation, while sparsely connected nodes suffer from insufficient spatial context. NRFormer+ inherits the density-adaptive spatial attention, which alternates between macro-scale and proximity-constrained attention, while introducing a critical modification to the query-key-value formulation that injects physics-aware radiation features.

\subsubsection{Physics-Enriched Query-Key Formulation}
A key architectural innovation of NRFormer+ is the redesigned input assignment for the spatial attention mechanism. In NRFormer, both the query/key and value matrices are derived from the fused temporal embedding $\mathbf{Z}_s^{(0)}$, meaning the spatial attention routing is determined by the same features that carry the information. In NRFormer+, we decouple these roles, setting $\mathbf{Q}_{s} = \mathbf{Z}_r$, $\mathbf{K}_{s} = \mathbf{Z}_r$, and $\mathbf{V}_{s} = \mathbf{Z}_f$, where $\mathbf{Z}_r \in \mathbb{R}^{N\times D}$ is the radiation feature representation from the temporal attention module (Section~\ref{sec:temporal_attention}), and $\mathbf{Z}_f \in \mathbb{R}^{N\times D}$ is the fused temporal embedding that aggregates radiation, physics, meteorological, location, temporal, and cross features.

This design has a clear physical interpretation. The radiation features in the query and key positions determine \emph{which stations should attend to each other} based on the similarity of their patterns, while the physics-enriched fused features in the value position determine \emph{what information is actually exchanged}. This separation ensures that spatial attention routing is grounded in observed radiation dynamics, while the exchanged information is enriched with physical context.

\subsubsection{Macro-Scale Spatial Correlation Modeling}
Following the physics-enriched formulation, we add learnable positional encoding and compute query, key, and value matrices $\mathbf{Q}_{g} = (\mathbf{Z}_r + \mathbf{P}_e) \mathbf{W}_{g}^Q$, $\mathbf{K}_{g} = (\mathbf{Z}_r + \mathbf{P}_e) \mathbf{W}_{g}^K$, and $\mathbf{V}_{g} = (\mathbf{Z}_f + \mathbf{P}_e) \mathbf{W}_{g}^V$. The macro-scale attention score is computed as
\begin{equation}
\begin{aligned}
\label{equ:global_A}
\mathbf{A}_{g} = \frac{\mathbf{Q}_{g} \cdot \mathbf{K}_{g}^{\top}}{\sqrt{D}},
\end{aligned}
\end{equation}
where $\mathbf{P}_e \in \mathbb{R}^{N\times D}$ is a learnable positional embedding matrix that preserves station identity information. The attention score matrix $\mathbf{A}_{g} \in \mathbb{R}^{N\times N}$ captures all-pair station relationships based on radiation pattern similarity.
The output is computed via
$\mathbf{H}_{g} = \text{Softmax}(\mathbf{A}_{g}) \cdot \mathbf{V}_{g}$,
followed by a feed-forward layer:
$\mathbf{H}^{\prime}_{g} = (\text{ReLU}(\mathbf{H}_{g}\cdot \mathbf{W}_1 + \mathbf{b}_1)) \cdot \mathbf{W}_2 + \mathbf{b}_2$.
The macro-scale attention provides additional spatial context for sparsely connected nodes, mitigating the under-smoothing problem.

\subsubsection{Proximity-Constrained Spatial Correlation Modeling}
For densely connected nodes, unconstrained all-pair attention may introduce excessive noise. We employ proximity-constrained attention using a graph mask derived from geographical distance. Let $\mathbf{A}$ denote the truncated adjacency matrix based on distance thresholds, we define:
\begin{equation}
\begin{aligned}
\mathbf{A}_{mask}[i,j] =
\begin{cases}
  \mathbf{A}_{l}[i,j], & \text{if}\ \mathbf{A}[i,j] > 0 \\
  -\infty, & \text{otherwise}
\end{cases}
\end{aligned}
\end{equation}
where $\mathbf{A}_{l}$ is the attention score matrix computed analogously to Eq.~\eqref{equ:global_A}. The constrained output is:
$\mathbf{H}_{l} = \text{Softmax}(\mathbf{A}_{mask}) \cdot \mathbf{V}_{l}$,
followed by a feed-forward layer:
$\mathbf{H}^{\prime}_{l} = (\text{ReLU}(\mathbf{H}_{l}\cdot \mathbf{W}_3 + \mathbf{b}_3)) \cdot \mathbf{W}_4 + \mathbf{b}_4$.
We alternately stack $L_s$ macro-scale and proximity-constrained spatial attention layers. The final spatial output is denoted as $\mathbf{Z}_s^{(L_s)} \in \mathbb{R}^{N\times D}$.

\subsection{Output Layer}
\label{sec:output_layer}
A key limitation of NRFormer's output layer is the two-way fusion of temporal and spatial embeddings, which may lose the original radiation signal after multiple layers of transformation. NRFormer+ addresses this through a three-way fusion strategy that preserves a direct radiation signal path:
\begin{equation}
\begin{aligned}
\label{equ:three_way_fusion}
\tilde{\mathbf{Y}}^{t+1:t+K}= \text{ReLU}\!\left((\mathbf{Z}_{r}||\mathbf{Z}_{f}||\mathbf{Z}_{s}^{(L_s)}) \cdot \mathbf{W}_i+\mathbf{b}_i\right)\cdot \mathbf{W}_o + \mathbf{b}_o,
\end{aligned}
\end{equation}
where $\mathbf{Z}_r \in \mathbb{R}^{N\times D}$ is the radiation feature embedding, $\mathbf{Z}_f \in \mathbb{R}^{N\times D}$ is the physics-enriched temporal fusion, $\mathbf{Z}_s^{(L_s)} \in \mathbb{R}^{N\times D}$ is the spatial attention output, $\mathbf{W}_i \in \mathbb{R}^{3D\times D_e}$ and $\mathbf{W}_o \in \mathbb{R}^{D_e\times K}$ are learnable weight matrices, and $\mathbf{b}_i, \mathbf{b}_o$ are bias parameters. Including $\mathbf{Z}_r$ directly in the fusion provides two benefits, (1) it creates a short-circuit path from the input radiation signal to the output, facilitating gradient flow during training, and (2) it allows the model to leverage the raw radiation representation even when the temporal and spatial transformations are insufficient.

The model output is denormalized using the reciprocal of the instance normalization:
\begin{equation}
\begin{aligned}
\hat{\mathbf{y}}_{i}= \sqrt{Var[\mathbf{x}_{i}] + \epsilon}\cdot \frac{\tilde{\mathbf{y}}_{i}-\beta_i}{\gamma_i} + \mathbb{E}[\mathbf{x}_{i}],
\end{aligned}
\end{equation}
where $\hat{\mathbf{y}}_{i} \in \mathbb{R}^{K}$ denotes the future prediction for station $i$.
The training objective minimizes the mean absolute error:
\begin{equation}
\mathcal{L}=\frac{1}{N}\sum_{i=1}^{N} |\hat{\mathbf{y}}_{i}-\mathbf{y}_{i}|,
\end{equation}
where $\mathbf{y}_{i}$ is the ground truth radiation level of station $i$. The MAE loss is chosen over MSE as it is more robust to the occasional extreme radiation spikes in the dataset, preventing the model from being disproportionately influenced by outlier events. The overall complexity of NRFormer+ is dominated by the spatial attention at $\mathcal{O}(L_s \cdot N^2 \cdot D)$, on par with NRFormer, with the added physics and context modules contributing only a modest, asymptotically subdominant overhead; a full per-module breakdown is given in Appendix~\ref{sec:complexity}.

\begin{table*}[ht]
\footnotesize
\renewcommand{\arraystretch}{0.85} % 调整行高
\setlength{\tabcolsep}{1.5pt} % 减小列间距
  \caption{Evaluations of NRFormer, NRFormer+, and baselines on two real-world datasets.
  We highlight the best accuracy in bold and underline the second-best accuracy.
  }
  \label{tab:overall_results}
  % \vspace{-0.3cm}
  \begin{tabular}{c|c|ccc|ccc|ccc|ccc|ccc}
    \toprule

    \multirow{2}{*}{\text { \textbf{Data} }} & \multirow{2}{*}{\text { \textbf{Models} }} & \multicolumn{3}{c|}{ \textbf{6th} (24 hours / 6 days)} & \multicolumn{3}{c|}{ \textbf{9th} (36 hours / 9 days)} & \multicolumn{3}{c|}{ \textbf{12th} (48 hours / 12 days)} & \multicolumn{3}{c|}{ \textbf{24th} (96 hours / 24 days)} & \multicolumn{3}{c}{ \textbf{sudden-change}}\\

    & & \text { MAE } & \text { RMSE } & \text { MAPE } & \text { MAE } & \text { RMSE } & \text { MAPE } & \text { MAE } & \text { RMSE } & \text { MAPE } & \text { MAE } & \text { RMSE } & \text { MAPE } & \text { MAE } & \text { RMSE } & \text { MAPE } \\
    
    \midrule

    \multirow{15}{*}{\rotatebox{90}{ Japan-4H }}
    & \text{ HA }            & 2.35 & 9.04 & 2.89\% & 2.42 & 9.45 & 2.94\% & 2.48 & 9.83 & 2.98\%  & 2.66 & 11.10 & 3.12\% & 3.47 & 8.24  & 4.69\% \\
    & \text{ LR }            & 2.08 & 6.74 & 2.77\% & 2.17 & 7.56 & 2.77\% & 2.29 & 8.04 & 2.99\%  & 2.43 & 9.89  & 3.00\% & 3.28 & 7.96  & 4.35\% \\
    & \text{ XGBoost }       & 2.17 & 8.49 & 2.63\% & 2.28 & 9.22 & 2.73\% & 2.35 & 9.27 & 2.80\%  & 2.52 & 10.47 & 2.95\% & 3.31 & 7.99  & 4.39\% \\

    & \text{ DCRNN }         & 1.96 & 6.75 & 2.59\% & 2.11 & 7.46 & 2.68\% & 2.26 & 7.95 & 2.86\%  & 2.36 & 8.73 & 2.95\% & 3.22 & 7.81 & 4.38\% \\
    & \text{ STID }          & 1.81 & 5.94 & 2.35\% & 1.89 & 6.38 & 2.44\% & 1.96 & 6.76 & 2.52\%  & 2.15 & 7.95 & 2.72\% & 3.08 & 7.80 & 4.06\% \\
    & \text{ DLinear }       & 1.96 & 6.36 & 2.61\% & 2.05 & 6.87 & 2.72\% & 2.12 & 7.26 & 2.79\%  & 2.29 & 8.49 & 2.97\% & 3.29 & 8.20 & 4.42\% \\
    & \text{ PatchTST }      & 1.85 & 6.44 & 2.42\% & 1.94 & 6.94 & 2.51\% & 2.01 & 7.35 & 2.58\%  & 2.18 & 8.53 & 2.76\% & 3.16 & 7.93 & 4.18\% \\
    & \text{ Koopa }         & 1.85 & 6.34 & 2.41\% & 1.93 & 6.83 & 2.49\% & 1.98 & 7.23 & 2.55\%  & 2.15 & 8.48 & 2.71\% & 3.18 & 7.87 & 4.24\% \\
    & \text{ iTransformer }  & \underline{1.73} & \underline{5.78} & \underline{2.27\%} & \underline{1.81} & \underline{6.24} & \underline{2.36\%} & \underline{1.88} & \underline{6.64} & 2.44\%  & \underline{2.07} & \underline{7.86} & \underline{2.63\%} & 3.06 & 7.75 & 4.08\% \\
    & \text{ TimesNet }      & 1.96 & 8.39 & 2.53\% & 2.01 & 8.67 & 2.58\% & 2.05 & 8.93 & 2.63\%  & 2.20 & 9.81 & 2.77\% & 3.32 & 8.16 & 4.43\% \\

    & \text{ StemGNN }       & 1.95 & 6.72 & 2.53\% & 2.12 & 7.43 & 2.62\% & 2.19 & 7.91 & 2.75\%  & 2.34 & 8.44 & 2.94\% & 3.16 & 7.65 & 4.27\% \\
    & \text{ GWN }           & 1.93 & 6.65 & 2.49\% & 2.02 & 7.15 & 2.60\% & 2.07 & 7.52 & 2.66\%  & 2.24 & 8.60 & 2.84\% & 3.17 & 8.51 & 4.20\% \\
    & \text{ LightCTS }      & 1.82 & 6.03 & 2.33\% & 1.90 & 6.49 & 2.42\% & 1.97 & 6.88 & 2.49\%  & 2.15 & 8.08 & 2.69\% & 3.16 & 8.00 & 4.18\% \\

    & \text{ NRFormer }      & 1.74 & 5.81 & 2.28\% & 1.82 & 6.27 & \underline{2.36\%} & 1.89 & 6.67 & \underline{2.43\%}  & \underline{2.07} & 7.92 & \underline{2.63\%} & \underline{2.86} & \underline{6.94} & \underline{3.47\%} \\
    & \text{ NRFormer+ }     & \textbf{1.72} & \textbf{5.69} & \textbf{2.26\%} & \textbf{1.79} & \textbf{6.06} & \textbf{2.25\%} & \textbf{1.83} & \textbf{6.48} & \textbf{2.33\%}  & \textbf{2.00} & \textbf{7.59} & \textbf{2.56\%} & \textbf{2.74} & \textbf{6.83} & \textbf{3.38\%} \\

    \midrule
    \multirow{15}{*}{\rotatebox{90}{Japan-1D }}
    & \text{ HA }            & 2.37 & 11.70 & 2.74\% & 2.48 & 12.74 & 2.85\% & 2.57 & 13.67 & 2.96\%  & 2.94 & 17.08 & 3.37\% & 4.16 & 10.62 & 5.48\% \\
    & \text{ LR }            & 2.03 & 6.77  & 2.69\% & 2.11 & 8.07  & 2.78\% & 2.18 & 8.91  & 2.78\%  & 2.41 & 11.48 & 3.17\% & 3.76 & 9.79  & 4.92\% \\
    & \text{ XGBoost }       & 2.25 & 10.25 & 2.61\% & 2.38 & 11.50 & 2.72\% & 2.46 & 12.59 & 2.81\%  & 2.69 & 15.91 & 3.02\% & 4.07 & 9.83  & 4.97\% \\

    & \text{ DCRNN }         & 2.19 & 9.56 & 2.62\% & 2.34 & 10.85 & 2.83\% & 2.42 & 9.12 & 2.89\%  & 2.61 & 12.76 & 3.29\% & 3.95 & 9.46  & 4.75\%   \\
    & \text{ STID }          & 2.02 & 6.93 & 2.63\% & 2.10 & 8.12  & 2.73\% & 2.16 & 8.99 & 2.81\%  & 2.42 & 11.57 & 3.15\% & 3.99 & 10.15 & 5.35\% \\
    & \text{ DLinear }       & 1.93 & 6.57 & 2.56\% & 2.04 & 8.04  & 2.70\% & 2.11 & 8.96 & 2.79\%  & 2.37 & 11.79 & 3.10\% & 4.05 & 9.98  & 5.43\% \\
    & \text{ PatchTST }      & \underline{1.83} & 6.69 & 2.41\% & \underline{1.94} & 8.07  & 2.53\% & 2.02 & 8.97 & 2.63\%  & 2.29 & 11.80 & 2.97\% & 3.83 & 9.98  & 4.97\% \\
    & \text{ Koopa }         & 1.89 & 6.78 & 2.47\% & 1.98 & 8.10  & 2.57\% & 2.07 & 8.99 & 2.67\%  & 2.32 & 11.69 & 2.98\% & 3.57 & 9.74  & 4.74\% \\
    & \text{ iTransformer }  & \underline{1.83} & \underline{6.42} & 2.43\% & \underline{1.94} & \underline{7.36}  & 2.55\% & 2.02 & \underline{8.13} & 2.64\%  & \underline{2.28} & \textbf{10.60} & 2.97\% & 3.42 & 9.91  & 4.51\% \\
    & \text{ TimesNet }      & 2.09 & 8.47 & 2.63\% & 2.15 & 9.16  & 2.70\% & 2.21 & 9.76 & 2.77\%  & 2.42 & 12.09 & 3.05\% & 3.77 & 10.54 & 4.80\% \\

    & \text{ StemGNN }       & 2.19 & 9.51 & 2.61\% & 2.34 & 10.79 & 2.88\% & 2.46 & 10.45 & 2.81\%  & 2.61 & 12.65 & 3.22\% & 3.82 & 9.36  & 4.56\%  \\
    & \text{ GWN }           & 2.15 & 9.24 & 2.61\% & 2.28 & 10.71 & 2.76\% & 2.37 & 11.24 & 2.88\%  & 2.59 & 12.60 & 3.21\% & 4.09 & 10.13 & 5.23\% \\
    & \text{ LightCTS }      & 1.95 & 6.60 & 2.57\% & 2.06 & 7.49  & 2.68\% & 2.13 & 8.18  & 2.78\%  & 2.37 & 11.50 & 3.09\% & 3.95 & 8.85  & 4.70\% \\

    & \text{ NRFormer }       & 1.84 & 6.58 & \underline{2.39\%} & \underline{1.94} & 7.99 & \underline{2.50\%} & \underline{2.01} & 8.92 & \underline{2.59\%}  & \underline{2.28} & 11.65 & \underline{2.93\%} & \underline{2.91} & \underline{8.46} & \underline{3.62\%} \\
    & \text{ NRFormer+ }      & \textbf{1.77} & \textbf{6.19} & \textbf{2.34\%} & \textbf{1.89} & \textbf{7.26} & \textbf{2.47\%} & \textbf{1.97} & \textbf{8.11} & \textbf{2.58\%}  & \textbf{2.22} & \underline{10.83} & \textbf{2.90\%} & \textbf{2.77} & \textbf{8.23} & \textbf{3.21\%} \\

    \bottomrule
  \end{tabular}
  % \vspace{-0.2cm}
\end{table*}

\section{Experiments}
\label{sec:Experiments}
% In this section, we introduce the experiment setup, overall performance, ablation study, and parameter sensitivity analysis.

\subsection{Experiments Setup}

\subsubsection{Datasets}
The experimental datasets for this predictive model include nuclear radiation data and meteorological data collected from the entire country of Japan. 
Table~\ref{tab:dataset_description} summarizes the statistics of the radiation and meteorological datasets.

\subsubsection{Evaluation Metrics}
We employed three widely used metrics, including Mean Absolute Error (MAE), Root Mean Squared Error (RMSE), and Mean Absolute Percentage Error (MAPE), for model evaluation. Lower values in MAE, RMSE, and MAPE indicate higher forecasting accuracy. 
Moreover, we follow \cite{zheng2015forecasting,yi2018deep,liang2023airformer} to discuss the errors on predicting sudden changes. We identified the top 10\% of samples in the test set with the greatest variation between observed and predicted values as sudden-change samples for prediction.

\subsubsection{Baseline}
We compare NRFormer+ against \textbf{13} baselines spanning classical statistics (HA~\cite{shao2022decoupled}, LR), gradient boosting (XGBoost~\cite{chen2016xgboost}), spatio-temporal graph neural networks (DCRNN~\cite{li2018diffusion}, GWN~\cite{wu2019graph}, StemGNN~\cite{cao2020spectral}, STID~\cite{shao2022spatial}, LightCTS~\cite{lai2023lightcts}), and recent time-series models (DLinear~\cite{zeng2023transformers}, PatchTST~\cite{nietime}, Koopa~\cite{liu2024koopa}, iTransformer~\cite{liu2024itransformer}, TimesNet~\cite{wutimesnet}). Detailed per-baseline descriptions are provided in Appendix~\ref{sec:baseline_details}.

\subsubsection{Implementation Details}
The dataset was partitioned into three distinct subsets, 60\% for training, 20\% for validation, and 20\% for testing. Our experiments are conducted on two different server configurations, Linux Centos with four RTX 3090 GPUs and Linux Ubuntu with two A800 GPUs.
In terms of implementation, our model is implemented using PyTorch, with the Adam optimizer chosen for optimization. We set the learning rate and the weight decay to 0.001 and 0.0001, respectively. The batch size is fixed to 32. The numbers of temporal and spatial attention layers $L_t$ and $L_s$ are selected via grid search per dataset (see Appendix~\ref{sec:parameter_sensitivity}); the spatial depth $L_s$ is fixed to 2 on both datasets, while $L_t$ is 4 on Japan-1D and 3 on Japan-4H. The head of the multi-head attention module is set to 4. Following standard practice for heavy-tailed targets, we model radiation in log space. Each dose rate $C$ is transformed by $\log(1+C)$ before standardisation, and predictions are mapped back with $\exp(\cdot)-1$ and clipped to the non-negative physical range at inference. This compresses the roughly two-order-of-magnitude dynamic range of radiation (about 40 to 7170\,nSv/h) so that the loss is not dominated by a few large values.

\subsection{Overall Performance}
Table~\ref{tab:overall_results} presents the overall performance of NRFormer, NRFormer+, and all baseline models on two real-world datasets with respect to MAE, RMSE, and MAPE.
NRFormer+ outperforms all baselines on MAE and MAPE across both datasets and all forecasting horizons, while also winning RMSE on the vast majority of (model, horizon) cells; the only exception is the longest 24-step horizon on Japan-1D, where iTransformer attains a marginally lower RMSE ($10.60$ vs $10.83$) -- though NRFormer+ still wins MAE ($2.22$ vs $2.28$) and MAPE ($2.90\%$ vs $2.97\%$) at that cell.
Specifically, NRFormer+ achieves notable improvements over the strongest Transformer baseline iTransformer in terms of the MAE metric across all forecasting horizons on both datasets.
Moreover, we can make the following observations, (1) All traditional methods (\ie HA, LR, and XGBoost) perform worse than deep learning models, as they fail to capture spatio-temporal patterns.
(2) STGNN-based approaches (\ie DCRNN, GWN) outperform HA, LR, and XGBoost by a large margin, indicating the superiority of modeling intricate spatio-temporal dependencies.
(3) STID, despite relying only on lightweight spatial and temporal identity embeddings, outperforms the graph-based DCRNN and GWN on Japan-1D across all horizons in MAE, suggesting that explicit graph convolution is not essential on this benchmark and that identity embeddings already capture the dominant spatio-temporal regularities.
(4) Among the deep-learning baselines, iTransformer achieves the strongest average performance by reinterpreting the Transformer for multivariate forecasting (variate-as-token attention), surpassing both classical and graph-based methods, though on Japan-1D it is matched almost exactly by PatchTST at the shorter horizons. A cluster of lightweight baselines (PatchTST, STID, Koopa, LightCTS, and DLinear) follows within a narrow band, the closest runner-up being PatchTST on Japan-1D and STID on Japan-4H, indicating that lightweight architectures are competitive on the radiation forecasting benchmark.

Moreover, the sudden-change experiments reveal that NRFormer+ shows substantial improvements across all datasets. By addressing spatio-temporal distribution imbalances, incorporating heterogeneous contextual factors, and leveraging physics-guided atmospheric modeling, NRFormer+ adapts more quickly to the non-stationary nature of radiation, providing more accurate predictions for sudden-change samples.

\begin{figure}[t]
  \centering
  \includegraphics[width=1\linewidth]{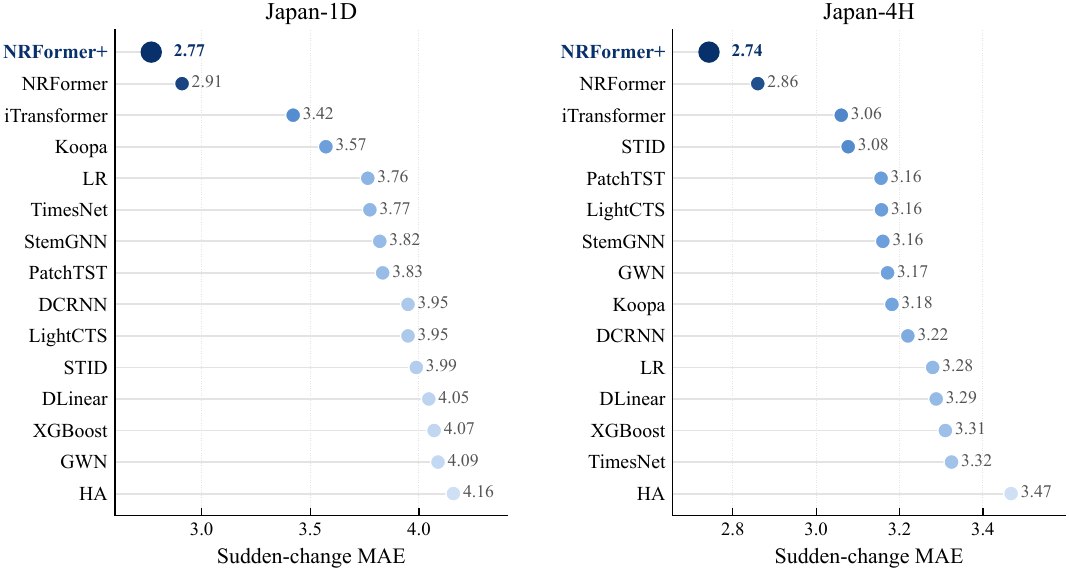}
  \caption{Sudden-change MAE leaderboard across all 15 evaluated models on Japan-1D (left) and Japan-4H (right), in ascending order.}
  \label{fig:sudden_comparison}
\end{figure}

\subsection{Sudden-Change Robustness}
\label{sec:sudden_change_robustness}
To further isolate sudden-change behavior from the aggregate table, Figure~\ref{fig:sudden_comparison} ranks all 15 evaluated models by their sudden-change MAE on the held-out test split. NRFormer+ ranks first on both datasets, and its closest competitor is the conference NRFormer. NRFormer+ improves upon the conference NRFormer by $4.87\%$ on Japan-1D ($2.7683$ versus $2.91$) and $4.09\%$ on Japan-4H ($2.7429$ versus $2.86$), confirming that the physics-guided atmospheric diffusion module and the enhanced encoders sharpen sudden-event tracking beyond the conference model. Relative to the strongest external (non-NRFormer) baseline iTransformer, the margin widens to $19.09\%$ on Japan-1D ($3.4212$) and $10.37\%$ on Japan-4H ($3.0602$), and the gap to all remaining baselines is uniformly larger. The systematic sudden-change advantage is consistent with the discussion in Appendix~\ref{sec:discussion_physics} that the physics module concentrates its contribution on the heavy-tailed sudden-change subset rather than on slowly varying periods.

\begin{table}[t]
\footnotesize
\renewcommand{\arraystretch}{0.85} % 调整行高
  \caption{Ablation study of NRFormer+ components on two datasets. Results are reported as single-step metrics at the 12th forecasting horizon.}
  \label{tab:ablation_nrformer_plus}
  \begin{tabular}{c|c|ccc}
    \toprule
    \text{Dataset} & \text{Variant} & \text{MAE} & \text{RMSE} & \text{MAPE} \\
    \midrule
    \multirow{6}{*}{Japan-4H}
    & \text{w/o Physics Module}          & 2.2147 & 11.7537 & 2.77\% \\
    & \text{w/o Enhanced Meteo Encoder}   & 2.0965 & 7.8886 & 2.62\% \\
    & \text{w/o Temporal Encoder}         & 2.1098 & 7.9394 & 2.66\% \\
    & \text{w/o RevIN}                    & 2.2227 & 8.0172 & 2.84\% \\
    & \text{w/o Log-space Target}         & 2.1033 & 8.0718 & 2.65\% \\
    & \textbf{NRFormer+ (Full)}           & \textbf{2.0021} & \textbf{7.8130} & \textbf{2.53\%} \\
    \midrule
    \multirow{6}{*}{Japan-1D}
    & \text{w/o Physics Module}          & 2.4167 & 11.5541 & 3.18\% \\
    & \text{w/o Enhanced Meteo Encoder}   & 2.2968 & 10.7409 & 2.97\% \\
    & \text{w/o Temporal Encoder} & 2.3398 & 10.4338 & 3.07\% \\
    & \text{w/o RevIN}                    & 2.4198 & 10.2917 & 3.09\% \\
    & \text{w/o Log-space Target}         & 2.2878 & 10.5755 & 2.94\% \\
    & \textbf{NRFormer+ (Full)}           & \textbf{2.2305} & \textbf{10.2591} & \textbf{2.83\%} \\
    \bottomrule
  \end{tabular}
\end{table}

For space, two supplementary analyses are deferred to the appendix. An ablation of the individual meteorological features (Appendix~\ref{sec:contextual_ablation}) shows that every covariate contributes, most strongly on the sudden-change subset, where removing all meteorology raises sudden-change MAE by up to $8.6\%$. A sensitivity analysis of the temporal and spatial attention depths $L_t$ and $L_s$ (Appendix~\ref{sec:parameter_sensitivity}) shows that all sweep points lie within $0.05$ MAE of the per-dataset optimum, confirming robustness to these hyperparameters.

\subsection{Ablation Study of NRFormer+ Components}
\label{sec:ablation_nrformer_plus}
To comprehensively evaluate the contribution of each component in NRFormer+, we conduct an extensive ablation study. The first three variants each remove one newly introduced architectural component. (1) \emph{w/o Physics Module} removes the Physics-Guided Atmospheric Diffusion Module, which estimates diffusion coefficients and computes spatial Laplacian features. (2) \emph{w/o Enhanced Meteo Encoder} replaces the dedicated wind and temperature pathway encoder with a simple concatenation followed by a single MLP. (3) \emph{w/o Temporal Encoder} removes the day-of-year seasonal embedding. Two further variants probe key training choices. (4) \emph{w/o RevIN} disables the instance-level reversible normalization in the temporal attention. (5) \emph{w/o Log-space Target} trains on linear-space radiation targets instead of the log-space target.

The results are reported in Table~\ref{tab:ablation_nrformer_plus}. Several important observations can be made. First, among the newly introduced architectural components, \emph{w/o Physics Module} exhibits the largest degradation on both datasets, and the physics prior is especially important on the heavy-tailed sudden-change subset, confirming the centrality of the atmospheric diffusion module; the reversible instance normalization (\emph{w/o RevIN}) and the log-space target (\emph{w/o Log-space Target}) are comparably impactful. Second, removing the enhanced meteorological encoder leads to notable accuracy drops, indicating that the dedicated wind and temperature pathways are more effective than naive feature concatenation. Third, the temporal encoder contributes consistent improvements, reflecting the strong seasonal periodicity in radiation levels. A finer-grained per-component visualization, together with a comparison of the diffusion module against a region-clustering alternative, is provided in Appendix~\ref{sec:extended_ablation}.

\subsection{Efficiency Analysis}
\label{sec:efficiency_analysis}
Beyond forecasting accuracy, computational efficiency is a critical consideration for practical deployment, especially given the large-scale nature of our nationwide monitoring network (3,627 stations). We conduct a systematic efficiency analysis comparing NRFormer+ against all baselines and NRFormer in terms of model size (\#Params), training time per epoch, and inference time per epoch over the full validation set. All timing measurements are wall-clock seconds collected on a single NVIDIA A800-SXM4-80GB GPU on the Japan-1D dataset. The results are summarized in Table~\ref{tab:efficiency}.

NRFormer+ has 343\,K parameters, only $22\%$ more than the conference NRFormer (282\,K) and within the same order of magnitude as the strongest Transformer baselines (iTransformer, 271\,K; TimesNet, 1.41\,M). Training takes 64.6\,s/epoch, $1.55\times$ the cost of the conference NRFormer (41.8\,s/epoch) and broadly comparable to iTransformer (42.8\,s/epoch) and TimesNet (41.4\,s/epoch), but $28\times$ faster than the heavyweight graph-convolutional baseline GWN (1849.5\,s/epoch). Inference takes 13.0\,s/epoch on the full validation set, on par with iTransformer (11.5\,s) and TimesNet (12.7\,s) and well within the latency budget of an operational system producing forecasts at 4-hour and daily intervals. Crucially, NRFormer+ also achieves the lowest 24-step MAE (2.22) across all methods, outperforming the strongest baselines NRFormer and iTransformer (both 2.28) by $2.6\%$ and the heavyweight GWN (2.59) by $14.3\%$. The favorable accuracy-efficiency trade-off demonstrates that the Physics-Guided Atmospheric Diffusion Module, the enhanced meteorological encoder, and the additional embedding branches in NRFormer+ deliver substantial accuracy gains at only a marginal computational premium.

\begin{table}[t]
\footnotesize
\renewcommand{\arraystretch}{0.85} % 调整行高
  \caption{Efficiency comparison of all deep-learning methods on the Japan-1D dataset.}
  \label{tab:efficiency}
  \begin{tabular}{l|cccc}
    \toprule
    \text{Model} & \text{\#Params (M)} & \makecell{\text{Train}\\\text{(s/epoch)}} & \makecell{\text{Inference}\\\text{(s/epoch)}} & \text{MAE} \\
    \midrule
    DCRNN        & 0.376  & 255.6   & 36.1   & 2.61 \\
    STID         & 0.143  & 22.2    & 6.6    & 2.42 \\
    DLinear      & 0.001  & 22.2    & 6.7    & 2.37 \\
    PatchTST     & 0.025  & 23.5    & 6.7    & 2.29 \\
    Koopa        & 5.700  & 22.7    & 6.7    & 2.32 \\
    StemGNN      & 4.878  & 304.18  & 43.39  & 2.61 \\
    GWN          & 0.486  & 1849.5  & 192.3  & 2.59 \\
    LightCTS     & 0.161  & 113.2   & 20.0   & 2.37 \\
    iTransformer & 0.271  & 42.8    & 11.5   & 2.28 \\
    TimesNet     & 1.409  & 41.4    & 12.7   & 2.42 \\
    \midrule
    NRFormer     & 0.282  & 41.8    & 5.6    & 2.28 \\
    \textbf{NRFormer+} & \textbf{0.343} & \textbf{64.6} & \textbf{13.0} & \textbf{2.22} \\
    \bottomrule
  \end{tabular}
\end{table}

\subsection{Case Study}
\label{sec:case_study}
To complement the aggregate metrics with a qualitative view of multi-step forecasting behavior, we conduct a case study at three named monitoring sites selected from the pool of 2,859 candidate stations falling within the three bounding boxes defined for Fukushima Prefecture, Tokyo Metropolis, and Shimane Prefecture (out of 3{,}627 total monitoring stations), Namiki Kindergarten in Fukushima Prefecture ($37.41^{\circ}\text{N}$, $140.36^{\circ}\text{E}$), the Ukishima Monitoring Post in Tokyo ($35.53^{\circ}\text{N}$, $139.77^{\circ}\text{E}$), and \=Oda High School in Shimane Prefecture ($35.19^{\circ}\text{N}$, $132.51^{\circ}\text{E}$). These three sites span a high-radiation legacy regime (Fukushima), a dense urban background regime (Tokyo), and a sparse coastal regime (Shimane), which together exercise the model across markedly different spatial and meteorological contexts. For each station we visualize the twelve-step-ahead forecasts of NRFormer+ alongside those of the strongest Transformer-based baseline, iTransformer, against the ground-truth measurements over a 33-day evaluation window from 2025-04-26 to 2025-05-29.

\begin{figure}[t]
\centering
\includegraphics[width=1\linewidth]{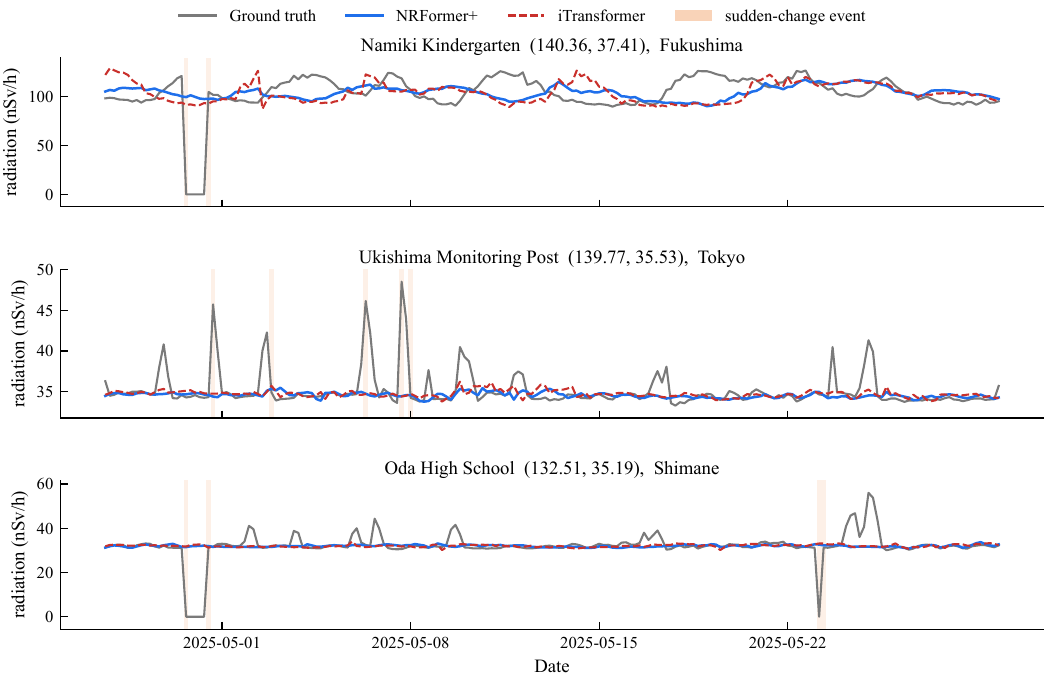}
\caption{Twelve-step-ahead forecasts of NRFormer+ and iTransformer against ground truth over a 33-day window (2025-04-26 to 2025-05-29) at three named stations spanning the high-amplitude Fukushima, urban Tokyo, and sparse Shimane regimes. Light vertical bands mark sudden-change events.}
\label{fig:case_study}
\end{figure}

Figure~\ref{fig:case_study} reports the resulting trajectories. At Namiki Kindergarten, where two pronounced meteorology-driven excursions occur within the plotted window, NRFormer+ reduces MAE from $15.33$ to $13.68$~nSv/h, a $10.76\%$ relative improvement, and visibly tracks both spikes with markedly smaller phase error than iTransformer. The Ukishima and \=Oda sites, which exhibit five and four smaller sudden-change events respectively, see more modest but still consistent advantages of $4.27\%$ ($1.102 \to 1.055$~nSv/h) and $2.41\%$ ($2.741 \to 2.675$~nSv/h) in MAE. Across all three panels NRFormer+ never trails iTransformer, indicating that the physics-guided atmospheric module and density-adaptive spatial attention introduced in NRFormer+ improve sudden-event tracking in the high-amplitude Fukushima regime without sacrificing accuracy in the quieter urban and sparse regimes. The mean MAE reduction across the three stations is $5.81\%$, consistent in sign with the aggregate sudden-change robustness gains reported in Section~\ref{sec:sudden_change_robustness}.

\section{Related Work}
\label{sec:related_work}

Spatio-temporal forecasting has been studied extensively in urban computing~\cite{jin2023spatio}, especially for traffic, where spatio-temporal graph networks jointly model spatial and temporal dependencies. Diffusion-convolutional and adaptive-graph models learn the sensor graph and propagate information over it~\cite{li2018diffusion,wu2019graph,wu2022autocts,cirstea2022towards,shao2022decoupled}, while lighter identity-embedding and efficient designs reach comparable accuracy at lower cost~\cite{shao2022spatial,lai2023lightcts}. In parallel, Transformers~\cite{vaswani2017attention} have been adapted to time-series forecasting through sparse or decomposition-based attention~\cite{zhou2021informer}, patching and channel independence~\cite{nietime}, variate-as-token attention~\cite{liu2024itransformer}, and multi-periodicity modeling~\cite{wutimesnet}, with a parallel line targeting non-stationarity via stationarization and reversible instance normalization~\cite{liu2022non,kim2021reversible}. These methods assume roughly uniform sensor coverage and do not model the physics-governed diffusion and meteorological context specific to radiation forecasting.

A complementary direction injects physical knowledge into neural models. Physics-informed neural networks add PDE-residual losses that constrain solutions to known laws~\cite{raissi2019physics}, with success in fluid dynamics and weather science~\cite{raissi2020hidden,beucler2021enforcing}, and neural operators learn PDE solution maps directly~\cite{li2021fourier}. Closest to us, AirPhyNet~\cite{hettige2024airphynet} encodes advection and diffusion as differential-equation operators on a sensor graph for air quality. Deep models have likewise advanced environmental forecasting more broadly~\cite{wang2022deep}, from latent-variable Transformers for PM$_{2.5}$~\cite{liang2023airformer} to Fourier-operator and multi-mesh-graph weather models~\cite{pathak2022fourcastnet,lam2023graphcast}. These approaches target dense, roughly uniform networks and a single pollutant, whereas radiation forecasting must additionally contend with extreme station-density imbalance, decay-driven multi-year non-stationarity, and sparse co-located meteorology. We therefore encode a discrete diffusion approximation as a soft architectural prior rather than a hard residual loss.

For nuclear radiation specifically, operational practice relies on Lagrangian dispersion solvers such as FLEXPART and HYSPLIT~\cite{stohl2005flexpart,stein2015hysplit} and on decision-support systems that couple dispersion with dose assessment~\cite{ehrhardt2000rodos}, while statistical work uses geostatistical interpolation and Bayesian source inversion~\cite{winiarek2012estimation}. Recent learning-based efforts remain local, for example combining random forests, Kalman filtering, and kriging near the Fukushima Daiichi site~\cite{sun2022spatial}, and carry no explicit transport prior. To our knowledge, ours is the first to cast nationwide, multi-station radiation forecasting as a graph-structured prediction problem with a learned transport prior.

\section{Conclusion}
\label{sec:Conclusion}
Nationwide nuclear radiation forecasting is safety-critical, yet its monitoring data are non-stationary in time, cluster into a small fraction of the territory in space, and co-evolve with meteorology through atmospheric transport that purely observational models overlook. Our central idea is to learn this physics rather than ignore it. We curate and release \emph{Japan-4H} and \emph{Japan-1D}, nationwide benchmarks drawn from over four years of recordings across 3{,}627 radiation stations paired with 228 meteorological stations, together with the full preprocessing pipeline. On top of these, we build \textbf{NRFormer+}, a spatio-temporal Transformer centered on a Physics-Guided Atmospheric Diffusion Module that reads local meteorology to estimate how readily radiation spreads, gauges how each station departs from its neighborhood, and reinjects this signal as an architectural prior alongside non-stationary temporal attention and density-adaptive spatial attention. Across thirteen baselines on both datasets and all horizons, NRFormer+ consistently improves accuracy, with its largest gains on sudden-change events, where it cuts error by up to 19.1\%. Future work will add calibrated probabilistic forecasts and anisotropic, wind-aware transport. By turning a sparse, uneven sensor network into reliable forecasts, NRFormer+ offers a practical instrument for emergency response and public health.

% \section*{Acknowledgments}
% This work was supported by the National Key R\&D Program of China (Grant No.2023YFF0725004), the General Research Fund of RGC under Grant HKUST16207222 and PolyU15207424, National Natural Science Foundation of China (Grant No.92370204), the Guangzhou Basic and Applied Basic Research Program under Grant No. 2024A04J3279, Education Bureau of Guangzhou Municipality, and CCF-DiDi GAIA Collaborative Research Funds. 

% \newpage
\bibliographystyle{IEEEtran}
\bibliography{sample-base}

@inproceedings{lyu2025nrformer,
  title={NRFormer: Nationwide Nuclear Radiation Forecasting with Spatio-Temporal Transformer},
  author={Lyu, Tengfei and Han, Jindong and Liu, Hao},
  booktitle={Proceedings of the 31st ACM SIGKDD Conference on Knowledge Discovery and Data Mining},
  year={2025},
  doi={10.1145/3711896.3737252}
}

@inproceedings{wutimesnet,
  title={TimesNet: Temporal 2D-Variation Modeling for General Time Series Analysis},
  author={Wu, Haixu and Hu, Tengge and Liu, Yong and Zhou, Hang and Wang, Jianmin and Long, Mingsheng},
  booktitle={The Eleventh International Conference on Learning Representations},
  year={2022}
}

@inproceedings{liang2023airformer,
  title={Airformer: Predicting nationwide air quality in china with transformers},
  author={Liang, Yuxuan and Xia, Yutong and Ke, Songyu and Wang, Yiwei and Wen, Qingsong and Zhang, Junbo and Zheng, Yu and Zimmermann, Roger},
  booktitle={Proceedings of the AAAI Conference on Artificial Intelligence},
  volume={37},
  number={12},
  pages={14329--14337},
  year={2023}
}

@inproceedings{hettige2024airphynet,
  title={{AirPhyNet}: Harnessing Physics-Guided Neural Networks for Air Quality Prediction},
  author={Hettige, Kethmi Hirushini and Ji, Jiahao and Xiang, Shili and Long, Cheng and Cong, Gao and Wang, Jingyuan},
  booktitle={The Twelfth International Conference on Learning Representations (ICLR)},
  year={2024}
}

@inproceedings{yi2018deep,
  title={Deep distributed fusion network for air quality prediction},
  author={Yi, Xiuwen and Zhang, Junbo and Wang, Zhaoyuan and Li, Tianrui and Zheng, Yu},
  booktitle={Proceedings of the 24th ACM SIGKDD international conference on knowledge discovery \& data mining},
  pages={965--973},
  year={2018}
}

@inproceedings{zheng2015forecasting,
  title={Forecasting fine-grained air quality based on big data},
  author={Zheng, Yu and Yi, Xiuwen and Li, Ming and Li, Ruiyuan and Shan, Zhangqing and Chang, Eric and Li, Tianrui},
  booktitle={Proceedings of the 21th ACM SIGKDD international conference on knowledge discovery and data mining},
  pages={2267--2276},
  year={2015}
}

@article{liu2024koopa,
  title={Koopa: Learning non-stationary time series dynamics with koopman predictors},
  author={Liu, Yong and Li, Chenyu and Wang, Jianmin and Long, Mingsheng},
  journal={Advances in Neural Information Processing Systems},
  volume={36},
  year={2024}
}

@inproceedings{nietime,
  title={A Time Series is Worth 64 Words: Long-term Forecasting with Transformers},
  author={Nie, Yuqi and Nguyen, Nam H and Sinthong, Phanwadee and Kalagnanam, Jayant},
  booktitle={The Eleventh International Conference on Learning Representations},
  year={2023}
}

@article{cao2020spectral,
  title={Spectral temporal graph neural network for multivariate time-series forecasting},
  author={Cao, Defu and Wang, Yujing and Duan, Juanyong and Zhang, Ce and Zhu, Xia and Huang, Congrui and Tong, Yunhai and Xu, Bixiong and Bai, Jing and Tong, Jie and others},
  journal={Advances in neural information processing systems},
  volume={33},
  pages={17766--17778},
  year={2020}
}

@inproceedings{zeng2023transformers,
  title={Are transformers effective for time series forecasting?},
  author={Zeng, Ailing and Chen, Muxi and Zhang, Lei and Xu, Qiang},
  booktitle={Proceedings of the AAAI conference on artificial intelligence},
  volume={37},
  number={9},
  pages={11121--11128},
  year={2023}
}

@article{lai2023lightcts,
  title={LightCTS: A Lightweight Framework for Correlated Time Series Forecasting},
  author={Lai, Zhichen and Zhang, Dalin and Li, Huan and Jensen, Christian S and Lu, Hua and Zhao, Yan},
  journal={Proceedings of the ACM on Management of Data},
  volume={1},
  number={2},
  pages={1--26},
  year={2023},
  publisher={ACM New York, NY, USA}
}

@article{vaswani2017attention,
  title={Attention is all you need},
  author={Vaswani, Ashish and Shazeer, Noam and Parmar, Niki and Uszkoreit, Jakob and Jones, Llion and Gomez, Aidan N and Kaiser, {\L}ukasz and Polosukhin, Illia},
  journal={Advances in neural information processing systems},
  volume={30},
  year={2017}
}

@article{liu2022non,
  title={Non-stationary transformers: Exploring the stationarity in time series forecasting},
  author={Liu, Yong and Wu, Haixu and Wang, Jianmin and Long, Mingsheng},
  journal={Advances in Neural Information Processing Systems},
  volume={35},
  pages={9881--9893},
  year={2022}
}

@article{dickey1979distribution,
  title={Distribution of the estimators for autoregressive time series with a unit root},
  author={Dickey, David A and Fuller, Wayne A},
  journal={Journal of the American Statistical Association},
  volume={74},
  number={366a},
  pages={427--431},
  year={1979},
  publisher={Taylor \& Francis}
}

@article{kpss1992testing,
  title={Testing the null hypothesis of stationarity against the alternative of a unit root: How sure are we that economic time series have a unit root?},
  author={Kwiatkowski, Denis and Phillips, Peter C B and Schmidt, Peter and Shin, Yongcheol},
  journal={Journal of Econometrics},
  volume={54},
  number={1-3},
  pages={159--178},
  year={1992},
  publisher={Elsevier}
}

@inproceedings{kim2021reversible,
  title={Reversible instance normalization for accurate time-series forecasting against distribution shift},
  author={Kim, Taesung and Kim, Jinhee and Tae, Yunwon and Park, Cheonbok and Choi, Jang-Ho and Choo, Jaegul},
  booktitle={International Conference on Learning Representations},
  year={2021}
}

@article{steinhauser2014comparison,
  title={Comparison of the Chernobyl and Fukushima nuclear accidents: a review of the environmental impacts},
  author={Steinhauser, Georg and Brandl, Alexander and Johnson, Thomas E},
  journal={Science of the total environment},
  volume={470},
  pages={800--817},
  year={2014},
  publisher={Elsevier}
}

@article{qiao2011predicting,
  title={Predicting the spread of nuclear radiation from the damaged Fukushima Nuclear Power Plant},
  author={Qiao, FangLi and Wang, GuanSuo and Zhao, Wei and Zhao, JieChen and Dai, DeJun and Song, YaJuan and Song, ZhenYa},
  journal={Chinese Science Bulletin},
  volume={56},
  pages={1890--1896},
  year={2011},
  publisher={Springer}
}

@article{jin2023spatio,
  title={Spatio-temporal graph neural networks for predictive learning in urban computing: A survey},
  author={Jin, Guangyin and Liang, Yuxuan and Fang, Yuchen and Shao, Zezhi and Huang, Jincai and Zhang, Junbo and Zheng, Yu},
  journal={IEEE Transactions on Knowledge and Data Engineering},
  year={2023},
  publisher={IEEE}
}

@article{wu2019graph,
  title={Graph wavenet for deep spatial-temporal graph modeling},
  author={Wu, Zonghan and Pan, Shirui and Long, Guodong and Jiang, Jing and Zhang, Chengqi},
  journal={arXiv preprint arXiv:1906.00121},
  year={2019}
}

@article{shao2022decoupled,
  title={Decoupled dynamic spatial-temporal graph neural network for traffic forecasting},
  author={Shao, Zezhi and Zhang, Zhao and Wei, Wei and Wang, Fei and Xu, Yongjun and Cao, Xin and Jensen, Christian S},
  journal={Proceedings of the VLDB Endowment},
  volume={15},
  number={11},
  pages={2733--2746},
  year={2022},
  publisher={VLDB Endowment}
}

@inproceedings{zhou2021informer,
  title={Informer: Beyond efficient transformer for long sequence time-series forecasting},
  author={Zhou, Haoyi and Zhang, Shanghang and Peng, Jieqi and Zhang, Shuai and Li, Jianxin and Xiong, Hui and Zhang, Wancai},
  booktitle={Proceedings of the AAAI conference on artificial intelligence},
  volume={35},
  number={12},
  pages={11106--11115},
  year={2021}
}

@article{sun2022spatial,
  title={Spatial and temporal prediction of radiation dose rates near Fukushima Daiichi Nuclear Power Plant},
  author={Sun, Dajie and Wainwright, Haruko and Suresh, Ishita and Seki, Akiyuki and Takemiya, Hiroshi and Saito, Kimiaki},
  journal={Journal of Environmental Radioactivity},
  volume={251},
  pages={106946},
  year={2022},
  publisher={Elsevier}
}

@inproceedings{shao2022spatial,
  title={Spatial-temporal identity: A simple yet effective baseline for multivariate time series forecasting},
  author={Shao, Zezhi and Zhang, Zhao and Wang, Fei and Wei, Wei and Xu, Yongjun},
  booktitle={Proceedings of the 31st ACM International Conference on Information and Knowledge Management},
  pages={4454--4458},
  year={2022}
}

@inproceedings{li2018diffusion,
  title={Diffusion Convolutional Recurrent Neural Network: Data-Driven Traffic Forecasting},
  author={Li, Yaguang and Yu, Rose and Shahabi, Cyrus and Liu, Yan},
  booktitle={International Conference on Learning Representations},
  year={2018}
}

@article{ager2019wildfire,
  title={The wildfire problem in areas contaminated by the Chernobyl disaster},
  author={Ager, Alan A and Lasko, Richard and Myroniuk, Viktor and Zibtsev, Sergiy and Day, Michelle A and Usenia, Uladzimir and Bogomolov, Vadym and Kovalets, Ivan and Evers, Cody R},
  journal={Science of the Total Environment},
  volume={696},
  pages={133954},
  year={2019},
  publisher={Elsevier}
}

@inproceedings{chen2016xgboost,
  title={Xgboost: A scalable tree boosting system},
  author={Chen, Tianqi and Guestrin, Carlos},
  booktitle={Proceedings of the 22nd ACM SIGKDD International Conference on Knowledge Discovery and Data Mining},
  pages={785--794},
  year={2016}
}

@article{han2024bigst,
  title={Bigst: Linear complexity spatio-temporal graph neural network for traffic forecasting on large-scale road networks},
  author={Han, Jindong and Zhang, Weijia and Liu, Hao and Tao, Tao and Tan, Naiqiang and Xiong, Hui},
  journal={Proceedings of the VLDB Endowment},
  volume={17},
  number={5},
  pages={1081--1090},
  year={2024},
  publisher={VLDB Endowment}
}

@article{stockie2011mathematics,
  title={The mathematics of atmospheric dispersion modeling},
  author={Stockie, John M},
  journal={SIAM Review},
  volume={53},
  number={2},
  pages={349--372},
  year={2011},
  publisher={SIAM}
}

@article{shuman2013emerging,
  title={The emerging field of signal processing on graphs: Extending high-dimensional data analysis to networks and other irregular domains},
  author={Shuman, David I and Narang, Sunil K and Frossard, Pascal and Ortega, Antonio and Vandergheynst, Pierre},
  journal={IEEE Signal Processing Magazine},
  volume={30},
  number={3},
  pages={83--98},
  year={2013},
  publisher={IEEE}
}

@article{raissi2019physics,
  title={Physics-informed neural networks: A deep learning framework for solving forward and inverse problems involving nonlinear partial differential equations},
  author={Raissi, Maziar and Perdikaris, Paris and Karniadakis, George E},
  journal={Journal of Computational Physics},
  volume={378},
  pages={686--707},
  year={2019},
  publisher={Elsevier}
}

@article{raissi2020hidden,
  title={Hidden fluid mechanics: Learning velocity and pressure fields from flow visualizations},
  author={Raissi, Maziar and Yazdani, Alireza and Karniadakis, George Em},
  journal={Science},
  volume={367},
  number={6481},
  pages={1026--1030},
  year={2020},
  publisher={AAAS}
}

@article{beucler2021enforcing,
  title={Enforcing analytic constraints in neural networks emulating physical systems},
  author={Beucler, Tom and Pritchard, Michael and Rasp, Stephan and Ott, Jordan and Baldi, Pierre and Gentine, Pierre},
  journal={Physical Review Letters},
  volume={126},
  number={9},
  pages={098302},
  year={2021},
  publisher={APS}
}

@article{stohl2005flexpart,
  title={Technical note: The Lagrangian particle dispersion model {FLEXPART} version 6.2},
  author={Stohl, Andreas and Forster, Carolien and Frank, Andreas and Seibert, Petra and Wotawa, Gerhard},
  journal={Atmospheric Chemistry and Physics},
  volume={5},
  number={9},
  pages={2461--2474},
  year={2005}
}

@article{stohl2012xenon,
  title={Xenon-133 and caesium-137 releases into the atmosphere from the {Fukushima Dai-ichi} nuclear power plant: determination of the source term, atmospheric dispersion, and deposition},
  author={Stohl, Andreas and Seibert, Petra and Wotawa, Gerhard and Arnold, Delia and Burkhart, John F and Eckhardt, Sabine and Tapia, C and Vargas, A and Yasunari, Teppei J},
  journal={Atmospheric Chemistry and Physics},
  volume={12},
  pages={2313--2343},
  year={2012}
}

@article{stein2015hysplit,
  title={NOAA's {HYSPLIT} atmospheric transport and dispersion modeling system},
  author={Stein, AF and Draxler, Roland R and Rolph, Glenn D and Stunder, Barbara JB and Cohen, MD and Ngan, Fong},
  journal={Bulletin of the American Meteorological Society},
  volume={96},
  number={12},
  pages={2059--2077},
  year={2015}
}

@article{ehrhardt2000rodos,
  title={The {RODOS} system: Decision support for off-site emergency management in {Europe}},
  author={Ehrhardt, J and Weis, A},
  journal={Radiation Protection Dosimetry},
  volume={91},
  number={1-3},
  pages={291--295},
  year={2000}
}

@article{winiarek2012estimation,
  title={Estimation of errors in the inverse modeling of accidental release of atmospheric pollutant: Application to the reconstruction of the cesium-137 and iodine-131 source terms from the {Fukushima Daiichi} power plant},
  author={Winiarek, Victor and Bocquet, Marc and Saunier, Olivier and Mathieu, Anne},
  journal={Journal of Geophysical Research: Atmospheres},
  volume={117},
  number={D5},
  year={2012}
}

@article{li2021fourier,
  title={Fourier neural operator for parametric partial differential equations},
  author={Li, Zongyi and Kovachki, Nikola and Azizzadenesheli, Kamyar and Liu, Burigede and Bhatt, Kaushik and Stuart, Andrew and Anandkumar, Anima},
  journal={arXiv preprint arXiv:2010.08895},
  year={2021}
}

@inproceedings{pathak2022fourcastnet,
  title={{FourCastNet}: A global data-driven high-resolution weather forecasting model},
  author={Pathak, Jaideep and Subramanian, Shashank and Harrington, Peter and Raja, Sanjeev and Chattopadhyay, Ashesh and Mardani, Morteza and Kurth, Thorsten and Hall, David and Li, Zongyi and Azizzadenesheli, Kamyar and others},
  booktitle={Proceedings of the Platform for Advanced Scientific Computing Conference},
  year={2022}
}

@article{lam2023graphcast,
  title={Learning skillful medium-range global weather forecasting},
  author={Lam, Remi and Sanchez-Gonzalez, Alvaro and Willson, Matthew and Wirnsberger, Peter and Fortunato, Meire and Alet, Ferran and Ravuri, Suman and Ewalds, Timo and Eaton-Rosen, Zach and Hu, Weihua and others},
  journal={Science},
  volume={382},
  number={6677},
  pages={1416--1421},
  year={2023},
  publisher={AAAS}
}

@inproceedings{liu2024itransformer,
  title={{iTransformer}: Inverted transformers are effective for time series forecasting},
  author={Liu, Yong and Hu, Tengge and Zhang, Haoran and Wu, Haixu and Wang, Shiyu and Ma, Lintao and Long, Mingsheng},
  booktitle={International Conference on Learning Representations},
  year={2024}
}

@article{qiu2024tfb,
  title={{TFB}: Towards Comprehensive and Fair Benchmarking of Time Series Forecasting Methods},
  author={Qiu, Xiangfei and Hu, Jilin and Zhou, Lekui and Wu, Xingjian and Du, Junyang and Zhang, Buang and Guo, Chenjuan and Zhou, Aoying and Jensen, Christian S. and Sheng, Zhenli and Yang, Bin},
  journal={Proceedings of the VLDB Endowment},
  volume={17},
  number={9},
  pages={2363--2377},
  year={2024},
  publisher={VLDB Endowment}
}

@article{wu2022autocts,
  title={{AutoCTS}: Automated Correlated Time Series Forecasting},
  author={Wu, Xinle and Zhang, Dalin and Guo, Chenjuan and He, Chaoyang and Yang, Bin and Jensen, Christian S.},
  journal={Proceedings of the VLDB Endowment},
  volume={15},
  number={4},
  pages={971--983},
  year={2022},
  publisher={VLDB Endowment}
}

@inproceedings{cirstea2022towards,
  title={Towards Spatio-Temporal Aware Traffic Time Series Forecasting},
  author={Cirstea, Razvan-Gabriel and Yang, Bin and Guo, Chenjuan and Kieu, Tung and Pan, Shirui},
  booktitle={2022 IEEE 38th International Conference on Data Engineering (ICDE)},
  pages={2900--2913},
  year={2022},
  organization={IEEE}
}

@article{wang2022deep,
  title={Deep learning for spatio-temporal data mining: A survey},
  author={Wang, Senzhang and Cao, Jiannong and Yu, Philip S},
  journal={IEEE Transactions on Knowledge and Data Engineering},
  volume={34},
  number={8},
  pages={3681--3700},
  year={2022},
  publisher={IEEE}
}
\vspace{-15 mm}
\begin{IEEEbiography}[{\includegraphics[width=1in,height=1.25in,clip,keepaspectratio]{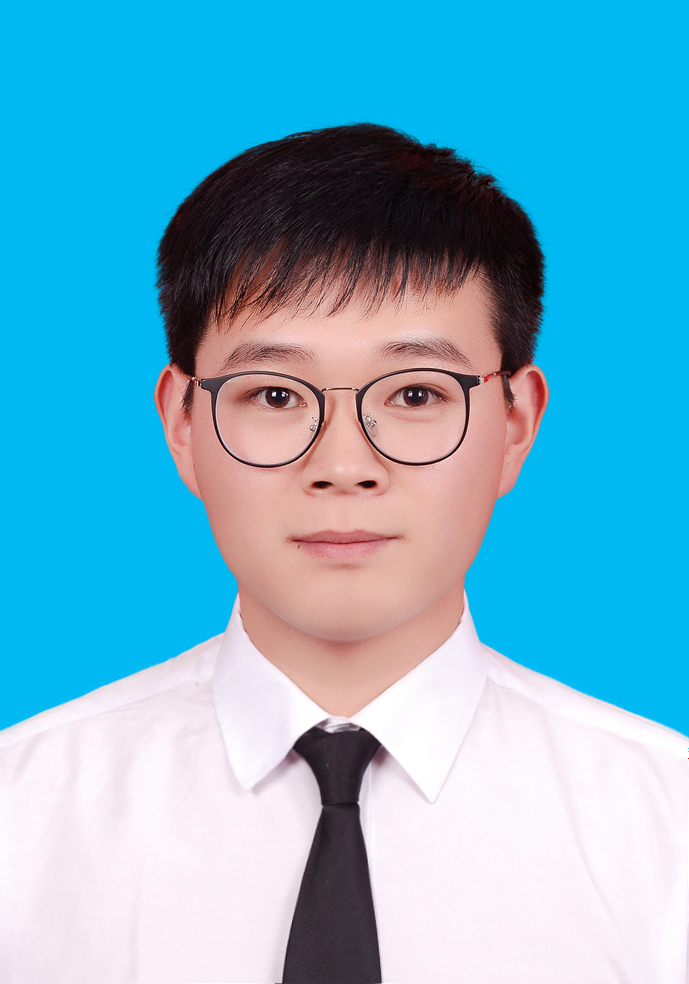}}]{Tengfei Lyu}
is a PhD student at the Thrust of Artificial Intelligence, The Hong Kong University
of Science and Technology (Guangzhou). He received the M.S. degree from the School of Computer Science and Engineering, Central South University, Changsha, China. His research interests are AI4Science, Spatiotemporal Data Mining, and Large Language Models including their theoretical foundations and applications. He has published several research papers in prestigious conferences and journals, such
as TCBB, SIGIR, IJCAI, and KDD.
\end{IEEEbiography}

\vspace{-15 mm}
\begin{IEEEbiography}[{\includegraphics[width=1in,height=1.25in,clip,keepaspectratio]{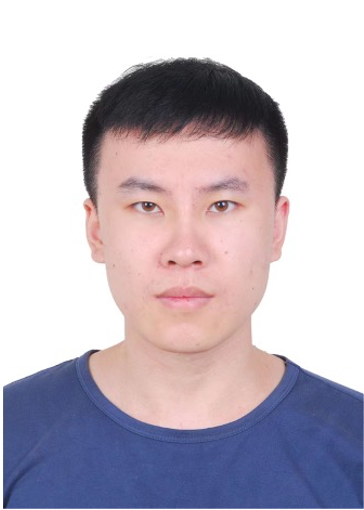}}]{Jindong Han}
is currently a Professor at the School of Artificial Intelligence, Shandong University. He received the Ph.D. degree from the the Hong Kong University of Science and Technology (HKUST) in 2025. His research focuses on data mining, urban intelligence, and AI for science. He have published over 30 papers in prestigious journals and conferences, such as IEEE TKDE, KDD, WWW, VLDB, NeurIPS, ICLR, and AAAI.
\end{IEEEbiography}

\vspace{-15 mm}
\begin{IEEEbiography}[{\includegraphics[width=1in,height=1.25in,clip,keepaspectratio]{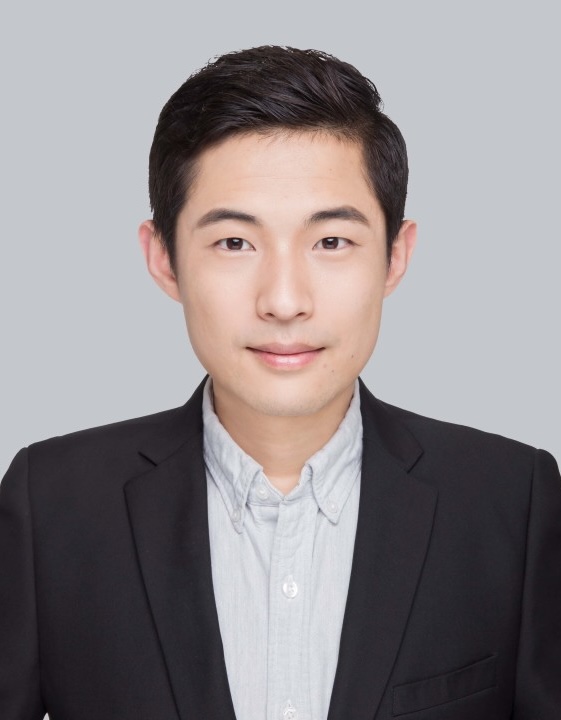}}]{Hao Liu}
received the BE degree from the South China University of Technology (SCUT), in 2012 and the PhD degree from the Hong Kong University of Science and Technology, in 2017. He is currently an Assistant Professor with the Thrust of Artificial Intelligence, The Hong Kong University
of Science and Technology (Guangzhou). Prior to that, he was a senior research scientist with Baidu Research and a postdoctoral fellow at HKUST. His general research interests are in data mining, machine learning, and Big Data management, with a special focus on mobile analytics and urban computing. He has published prolifically in refereed journals and conference proceedings, such as TKDE, KDD, ICML, NeurIPS, SIGIR, WWW, AAAI, and IJCAI.
\end{IEEEbiography}

\clearpage
\appendices

\section{Discussion}
\label{sec:discussion}

\subsection{Why Physics-Guided Modeling Helps}
\label{sec:discussion_physics}

\begin{figure*}[t]
  \centering
  \includegraphics[width=1\linewidth]{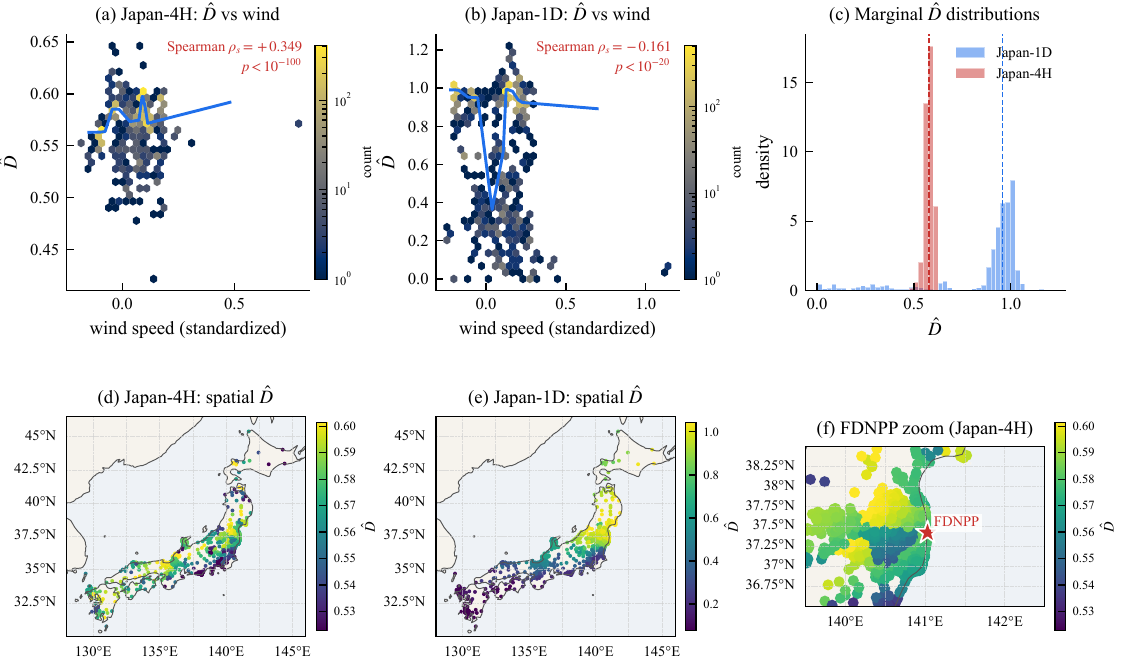}
  \caption{Physical interpretability of the learned diffusion coefficient $D_i$. (a, b)~Hexbin scatter of $D_i$ versus standardised wind speed on Japan-4H and Japan-1D, overlaid with running-median curves; the Spearman correlations are $\rho_s{=}{+}0.349$ ($p{=}2.6{\times}10^{-104}$) on Japan-4H and $\rho_s{=}{-}0.161$ ($p{=}1.7{\times}10^{-22}$) on Japan-1D. (c)~Overlaid histograms of $D_i$ on the two datasets. (d, e)~Geographic distribution of $D_i$ on Japan-4H and Japan-1D. (f)~Zoom-in around the Fukushima Daiichi site (red star) on Japan-4H, illustrating the local diffusion structure inferred by the model.}
  \label{fig:physics_interpretability}
\end{figure*}

The ablation in Section~\ref{sec:ablation_nrformer_plus} establishes that removing the Physics-Guided Atmospheric Diffusion Module produces one of the two largest accuracy drops among NRFormer+ components, alongside reversible instance normalization (a relative MAE increase of $\Delta\text{MAE}{=}{+}10.62\%$ on Japan-4H and ${+}8.35\%$ on Japan-1D at the 12-step horizon), and that the gain is concentrated in regimes where physics is most informative. Stratifying the test set by event type, the diffusion module contributes the bulk of its improvement on the heavy-tailed sudden-change subset identified in Finding T3, where deviations from the spatial mean are large and the Laplacian term $C_i^{(P)} - \bar{C}^{(P)}$ carries strong signal. Geographically, we expect the gains to be largest at sites that are spatially isolated (the sparse-neighbor regions identified in Finding S1), where data-driven smoothing has little to anchor on but the wind-conditioned diffusion coefficient can provide an external prior. Inspecting the learned $D_i$ values (Figure~\ref{fig:physics_interpretability}) yields a physically coherent picture. On Japan-4H, $D_i$ increases with mean wind speed (Spearman $\rho_s{=}{+}0.349$, $p{<}10^{-100}$), in line with the advection-diffusion intuition that stronger winds enhance horizontal mixing, and the model recovers this trend without direct supervision. On Japan-4H this learned coefficient runs opposite to the empirical isotropic-fit slope $\hat{D}$ in Finding M2, which instead declines with wind because a wind-blind operator captures less of the transport as the flow turns anisotropic. A learned wind-modulated operator is designed precisely to absorb the anisotropic transport that the isotropic fit loses. On Japan-1D the learned $D_i$ instead shares the sign of $\hat{D}$, both declining with wind, consistent with the directional-advection reading discussed below. The case study further illustrates this mechanism. At the high-amplitude Namiki site near Fukushima, NRFormer+ tracks two pronounced meteorology-driven excursions that iTransformer follows with markedly larger phase error, while even at the sparse Shimane site, where the gain is smallest, it never trails iTransformer. Because the case study compares the full model against iTransformer rather than against the w/o-Physics variant, it illustrates rather than isolates the contribution of the diffusion module.

Figure~\ref{fig:physics_interpretability} provides direct visual evidence for the physical interpretability of the diffusion module. On Japan-4H (panel~a), $D_i$ ranges between $0.42$ and $0.65$ and increases with wind speed, in line with the advection--diffusion intuition that stronger winds enhance horizontal mixing; on Japan-1D (panel~b), the model learns a broader, partially inverted distribution (range $0.00$--$1.22$, $\rho_s{=}{-}0.161$), and this negative correlation at the coarser daily scale is not fully explained by the isotropic diffusion picture, possibly reflecting wind acting more as a directional advection driver, a point we do not resolve here. The geographic maps (panels~d--e) and the Fukushima zoom-in (panel~f) suggest that elevated $D_i$ values concentrate in coastal and topographically open regions, while inland mountainous corridors exhibit lower diffusivity, a spatial pattern qualitatively consistent with terrain-modulated mixing that emerges without direct supervision of $D_i$.

\subsection{From Advection--Diffusion to Architecture: Design Choices and Trade-offs}
\label{sec:discussion_design}

A natural question is whether NRFormer+ would benefit from a stricter physics-informed formulation. We deliberately chose three soft-design points and discuss each in turn. First, we encode the physics triplet $[C, D, \nabla^2 C]$ as an \emph{auxiliary feature pathway} rather than as a PDE-residual loss in the style of canonical PINNs~\cite{raissi2019physics}. PINN-style residual losses are most effective when the governing equation is locally exact and observations are dense in space-time; nationwide radiation transport, in contrast, is contaminated by unobserved source terms, deposition, and irregular sampling, so an exact residual loss would propagate model-misspecification error directly into the gradient. Treating physics as a representation prior instead allows the model to defer to data wherever the diffusion equation is locally inadequate, which we argue is the dominant regime in this application. Second, the Laplacian is approximated as deviation from the spatial mean, $\nabla^2 C_i \approx C_i^{(P)} - \bar C^{(P)}$, rather than as a graph Laplacian $\sum_j W_{ij}(C_i - C_j)$. The mean-deviation form is parameter-free, robust to the extreme density imbalance of the monitoring network (Finding S1), and avoids re-introducing the over-smoothing/under-smoothing pathology that motivated density-adaptive attention in the first place; a $k$-NN graph Laplacian, by contrast, would inherit the same density bias. The cost is anisotropy, since a single scalar deviation cannot distinguish upwind from downwind neighbors, which we view as the leading direction for future tightening (Appendix~\ref{sec:discussion_future}). Third, we do not introduce an explicit advection operator. Advection is instead absorbed by the density-adaptive spatial attention, whose physics-enriched query/key formulation (Section~\ref{sec:spatial_attention}) routes information by learned radiation similarity rather than by a hand-coded wind kernel. This is consistent with the multi-hour lagged-correlation structure documented in Finding M1, where an instantaneous $\mathbf{u}\cdot\nabla C$ term would mis-specify the time scale, whereas attention can flexibly aggregate over the empirical $\sim$16\,h lag. Together these choices articulate a deliberate hybrid, applying hard physics where the equation is locally faithful, such as the positivity of $D$, and soft physics elsewhere.

\subsection{Limitations}
\label{sec:discussion_limitations}

We highlight six limitations together with their practical implications. \emph{(i) Single-country evaluation.} All results are obtained on Japan, whose monitoring network has both an exceptionally high baseline density and a strong post-Fukushima skew (Finding S1); transferring NRFormer+ to networks that are sparser overall (e.g., national-grid scale in continental Europe or North America) would stress both the imbalance-aware attention and the Laplacian approximation, and is likely to require re-calibration of the proximity threshold. \emph{(ii) Isotropic Laplacian.} The mean-deviation operator is rotation-invariant; under strongly directional transport (\eg jet-stream advection of Fukushima discharge plumes), it under-represents anisotropy, which Finding M2 indicates is the regime in which residual transport variance is largest. \emph{(iii) Static spatial graph.} Geographic proximity is fixed at training time, so the model cannot dynamically rewire connections when synoptic-scale circulation patterns redirect transport pathways across days. \emph{(iv) Meteorological interpolation.} The mapping from 228 NOAA-ISD stations to 3,627 radiation sites introduces an irreducible interpolation error whose magnitude scales with local meteorological gradients and station-to-station distance (ECDF reported in Figure~\ref{fig:meteo_coupling}(a)); in countries with sparser weather networks this error would compound and likely dominate physics-module gains. \emph{(v) Deterministic point forecasts.} NRFormer+ outputs a single trajectory and provides no calibrated uncertainty, which is a non-trivial limitation for safety-critical decisions such as evacuation timing or agricultural advisories where the cost of false negatives is asymmetric. \emph{(vi) Quadratic spatial attention.} Macro-scale attention is $\mathcal{O}(N^2)$ in stations; at $N{=}3{,}627$ this remains tractable on a single A800 GPU (Section~\ref{sec:efficiency_analysis}), but scaling to continental networks ($N{\gtrsim}10^4$) would require sparse or linearised attention, the integration of which with the radiation-guided query/key routing of Section~\ref{sec:spatial_attention} is non-trivial.

\subsection{Practical Implications and Broader Applicability}
\label{sec:discussion_practical}

NRFormer+ is best understood as complementary to, rather than a replacement for, established Lagrangian dispersion models such as FLEXPART~\cite{stohl2005flexpart} and HYSPLIT~\cite{stein2015hysplit}. Lagrangian models excel at \emph{source-attributed scenario simulation}. Given a hypothesised release, they trace particle trajectories through high-resolution numerical weather fields and produce physically auditable forecasts, at the cost of requiring an accurate source term and tens of minutes to hours of compute. NRFormer+ occupies the orthogonal niche of \emph{source-agnostic nationwide nowcasting}. It consumes the live monitoring stream and produces forecasts for all 3{,}627 stations in seconds per forward pass (Section~\ref{sec:efficiency_analysis}), with no need for source-term hypotheses. In an operational nuclear-safety stack the two approaches could be chained, with NRFormer+ continuously screening the network for anomalies and identifying stations likely to exceed regulatory thresholds, after which a Lagrangian model can be invoked for forensic source attribution and dose assessment. Beyond radiation, the architectural recipe of density-adaptive attention coupled with a soft physics-triplet pathway is in principle applicable to other environmental scalars governed by advection--diffusion dynamics on irregular sensor networks, such as ground-level $\text{PM}_{2.5}$, $\text{NO}_2$, and ground-water tracer transport, though transfer to a new network would require re-calibration and is left to future work.
\subsection{Future Directions}
\label{sec:discussion_future}

Five directions, ordered by methodological reach, follow naturally from the limitations above. \emph{Probabilistic forecasting.} Replacing the deterministic output head with quantile regression or a deep ensemble would provide calibrated predictive intervals essential for risk-informed decision making, and would couple naturally with the existing three-way fusion. \emph{Anisotropic, dynamic physics.} The mean-deviation Laplacian can be upgraded to a wind-modulated graph Laplacian whose edge weights depend on the projection of the wind vector onto the inter-station displacement, and the full advection term $-\mathbf{u}\cdot\nabla C$ can be parameterized as a learnable directional operator, jointly relaxing limitations \emph{(ii)}--\emph{(iii)}. \emph{Dynamic graph learning.} Replacing the static $k$-NN graph with a graph-structure-learning module conditioned on synoptic wind fields would let the receptive field follow transport pathways rather than geography. \emph{Cross-country generalization.} Pre-training on Japan-4H/Japan-1D and fine-tuning on European or North American monitoring networks would test the robustness of the design choices in Appendix~\ref{sec:discussion_design} under different density regimes, and would clarify which components are dataset-specific. \emph{Environmental foundation models.} The convergence of large-scale weather foundation models (FourCastNet~\cite{pathak2022fourcastnet}, GraphCast~\cite{lam2023graphcast}) suggests a similar trajectory for environmental scalar transport, pre-training on a heterogeneous corpus of radiation, air-quality, and tracer-dispersion time series, with NRFormer+'s physics-triplet pathway serving as the structural prior that disciplines transfer across domains.

\subsection{Relationship to the Conference Version}
\label{sec:relation-to-conference}

This paper substantially extends our KDD~2025 conference paper, NRFormer~\cite{lyu2025nrformer}. The differences are concentrated along five complementary axes that together substantially exceed the novelty expected of a journal extension.

\noindent\emph{(1) Reframed challenges and physics-grounded methodology.}
The conference version organised the problem around three data-driven challenges (non-stationary temporal patterns, imbalanced spatial distribution, and heterogeneous contextual factors) and addressed them via non-stationary temporal attention, imbalance-aware spatial attention, and a radiation-propagation prompting strategy. The journal version retains the temporal and spatial challenges but \emph{merges} the third with the previously implicit gap of missing physics priors into a single, unified challenge of \emph{physics-grounded modeling of heterogeneous context}. This reframing motivates a new, central architectural component, the Physics-Guided Atmospheric Diffusion Module, which estimates a meteorology-conditioned diffusion coefficient $D$, approximates the spatial Laplacian $\nabla^{2}C$ through a parameter-free mean-field surrogate based on each station's deviation from the network mean, and injects the physics triplet $[C,D,\nabla^{2}C]$ into the Transformer as an inductive bias derived from Eq.~\eqref{eq:adv-diff-intro}. This module is absent from the conference version.

\noindent\emph{(2) Architectural enhancements.}
Beyond the diffusion module, NRFormer+ introduces dedicated wind and thermal pathways in the meteorological encoder (Section~\ref{sec:meteo_encoder}), a day-of-year seasonal embedding that supersedes the three independent time-of-day / day-of-week / month embeddings of NRFormer (Section~\ref{sec:temporal_encoder}), a deeper location encoder and a radiation--location cross-feature encoder (Sections~\ref{sec:location_encoder}--\ref{sec:cross_features}), physics-enriched query/key routing in the density-adaptive spatial attention (Section~\ref{sec:spatial_attention}), and a three-way output fusion (radiation + temporal + spatial; Section~\ref{sec:output_layer}) that replaces the two-way fusion in NRFormer.

\noindent\emph{(3) Dataset extension.}
We extend the temporal coverage from three to over four years (March~2021--May~2025), refine the station set from $3{,}841$ to $3{,}627$ via a stricter, four-stage quality-control pipeline, expand the meteorological side to $228$ NOAA-ISD stations, and release the resulting Japan-4H ($9{,}222$ timesteps) and Japan-1D ($1{,}537$ timesteps) benchmarks together with all preprocessing code.

\noindent\emph{(4) Data engineering and analysis.}
A new Section~\ref{sec:data_description_and_analysis} documents the construction pipeline and reports a battery of quantitative analyses, including non-stationarity (ADF/KPSS, Findings T1--T3), spatial heterogeneity (Moran's $I$, semivariogram; Findings S1--S2), and meteorology--radiation coupling (lagged cross-correlation, wind-stratified diffusion regression; Findings M1--M2). Each finding is explicitly mapped to a corresponding architectural component of NRFormer+, so that the model is traceable to verified properties of the data rather than to design analogy.

\noindent\emph{(5) Experiments and ablations.}
We re-run all baselines on the extended benchmark, add ablations that isolate every new component of NRFormer+ (Section~\ref{sec:ablation_nrformer_plus}), report an efficiency comparison (Section~\ref{sec:efficiency_analysis}), and present a case study across high-, low-, and sparse-radiation regimes (Section~\ref{sec:case_study}). The discussion further provides a methodological reflection on the soft-physics design choices that was absent from the conference paper.

\section{Additional Experiments}
\label{sec:additional_experiments}

\subsection{Baseline Details}
\label{sec:baseline_details}
We compare the proposed NRFormer+ with 13 baseline methods.
\begin{itemize}
\item HA~\cite{shao2022decoupled}: predicts radiation levels by averaging historical readings for the corresponding periods.
\item XGBoost~\cite{chen2016xgboost}: predicts radiation levels with an ensemble of gradient-boosted decision trees.
\item LR: a linear regression model relating input variables to the output.
\item DCRNN~\cite{li2018diffusion}: a spatio-temporal model capturing spatial dependencies via bidirectional random walks and temporal dependencies via an encoder-decoder with scheduled sampling.
\item GWN~\cite{wu2019graph}: a spatio-temporal graph framework that learns an adaptive graph and integrates diffusion graph convolution with dilated causal convolution.
\item STID~\cite{shao2022spatial}: a simple model that augments historical time series with spatial-temporal identity embeddings.
\item DLinear~\cite{zeng2023transformers}: a one-layer linear forecaster that rivals complex Transformer models on long-term forecasting.
\item StemGNN~\cite{cao2020spectral}: captures inter-series correlations and temporal dependencies in the spectral domain.
\item LightCTS~\cite{lai2023lightcts}: an efficient, lightweight framework for correlated time-series forecasting.
\item PatchTST~\cite{nietime}: a Transformer using subseries patching and channel independence for multivariate forecasting.
\item Koopa~\cite{liu2024koopa}: forecasts non-stationary series via Koopman theory, disentangling time-variant and time-invariant dynamics.
\item iTransformer~\cite{liu2024itransformer}: treats each variate as a token and applies inverted self-attention across variates.
\item TimesNet~\cite{wutimesnet}: detects multi-periodicity via FFT and reshapes 1D series into 2D tensors processed by an inception-style backbone.
\end{itemize}

\subsection{Contextual Feature Ablation}
\label{sec:contextual_ablation}
\begin{figure}[t]
  \centering
  \includegraphics[width=1\linewidth]{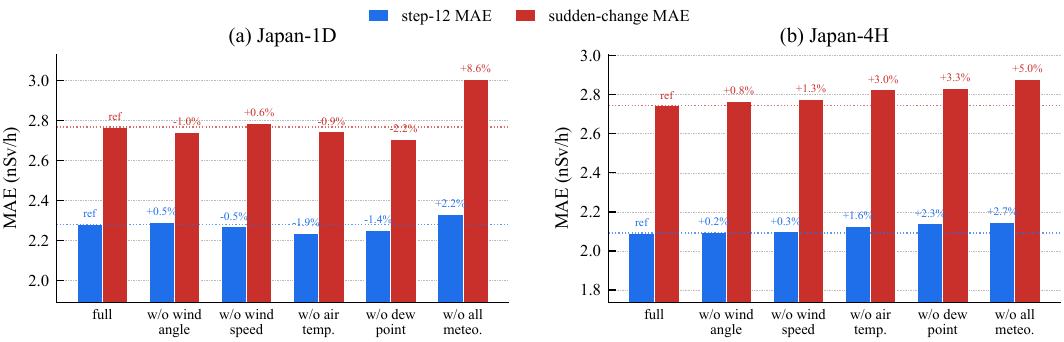}
  \vspace{-0.6cm}
  \caption{Ablation study of each contextual feature. }
  \label{fig:prompt_sensitivity}
  \vspace{-0.6cm}
\end{figure}

We examine the utility of the contextual features we provide to the prompting module. noPrompt means that we remove all the contextual features from our framework.
Others (\eg Wind Speed) indicate using only one contextual feature.
The results on MAE and RMSE are presented in Figure~\ref{fig:prompt_sensitivity}. In the absence of any contextual features, we separately add one feature at each time to assess its contribution. Here \emph{noPrompt} indicates the performance of NRFormer+ without utilizing any additional contextual features.
We can observe that incorporating each feature consistently improves the model performance beyond the \emph{noPrompt}. In summary, the contextual features have a significant impact on nuclear radiation, and taking them into account is beneficial for forecasting.

Concretely, on Japan-4H every meteorological feature removal degrades both step-12 MAE and sudden-change MAE. Removing the dew-point feature raises step-12 MAE by ${+}2.28\%$ and sudden MAE by ${+}3.32\%$; removing wind speed by ${+}0.27\%$ and ${+}1.26\%$; removing all meteorology by ${+}2.70\%$ and ${+}4.96\%$. On Japan-1D the individual single-feature removals are within run-to-run noise on step-12 MAE, but the sudden-change subset is markedly more sensitive. Removing all meteorology raises sudden MAE by ${+}8.62\%$, confirming that the meteorological prompts contribute primarily where weather drives episodic radiation fluctuations (cf.~Section~\ref{sec:meteo_coupling}).

\subsection{Extended Component Ablation}
\label{sec:extended_ablation}

\begin{figure}[t]
  \centering
  \includegraphics[width=1\linewidth]{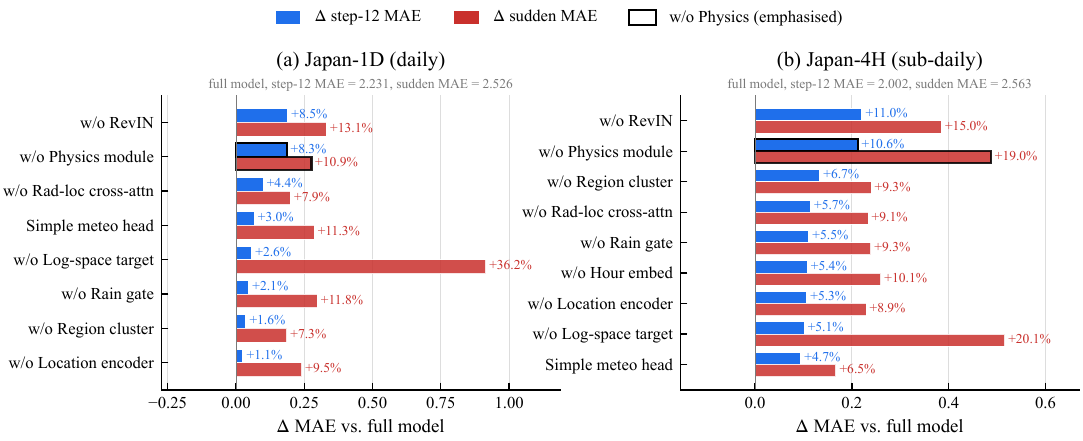}
  \caption{Per-component ablation of NRFormer+.}
  \label{fig:component_ablation}
\end{figure}

Figure~\ref{fig:component_ablation} visualizes the relative MAE deltas for all single-component ablations. Two variants dominate the 12-step accuracy hit on both datasets, \emph{w/o Physics} (${+}8.35\%$ on Japan-1D, ${+}10.62\%$ on Japan-4H) and \emph{w/o RevIN} (${+}8.49\%$, ${+}11.02\%$), confirming the centrality of the physics module and instance normalization. On the heavy-tailed sudden-change subset, both the physics prior and the log-space target matter most. Removing the physics module raises sudden-change MAE by ${+}19.03\%$ on Japan-4H, and replacing the log-space target by linear space (\emph{w/o Log-space Target}) raises it by ${+}36.17\%$ on Japan-1D and ${+}20.11\%$ on Japan-4H. The remaining single-component ablations (enhanced meteorological encoder, temporal encoder, location encoder, and rad-loc cross features) produce smaller but consistently positive drops.

\begin{figure}[t]
  \centering
  \includegraphics[width=1\linewidth]{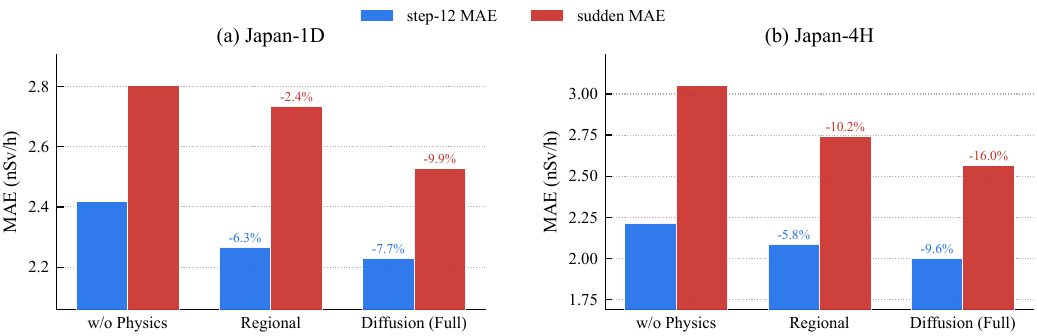}
  \caption{Soft-physics inductive bias versus a regional-clustering alternative. Three NRFormer+ configurations, w/o Physics, Regional (the diffusion module replaced by a region-cluster bias), and Diffusion (Full), evaluated on Japan-1D and Japan-4H in step-12 MAE and sudden-change MAE. Both physics-aware variants improve over w/o Physics, with the learned diffusion operator recovering the most on the heavy-tailed sudden-change subset.}
  \label{fig:physics_vs_regional}
\end{figure}

Figure~\ref{fig:physics_vs_regional} probes whether the gains attributed to the physics module reflect the diffusion parameterization specifically or a more generic meteorology-aware inductive bias. A generic Regional variant, which replaces $D_i$ with a region-cluster bias term, recovers part of the accuracy lost to \emph{w/o Physics}, but the learned diffusion operator recovers substantially more, and the gap is largest on the heavy-tailed sudden-change subset where physics is most informative. On Japan-4H the diffusion module cuts sudden-change MAE by $-16.0\%$ relative to \emph{w/o Physics} versus $-10.2\%$ for the regional bias, and on Japan-1D by $-9.9\%$ versus $-2.4\%$; the step-12 MAE shows the same ordering. This indicates that the wind-modulated diffusion parameterization contributes beyond a coarse meteorology-aware bias and justifies retaining the learned operator over a region-clustering surrogate, consistent with the soft-physics design rationale in Appendix~\ref{sec:discussion_design}.

\subsection{Parameter Sensitivity Analysis}
\label{sec:parameter_sensitivity}
\begin{figure}[t]
  \centering
  \includegraphics[width=1\linewidth]{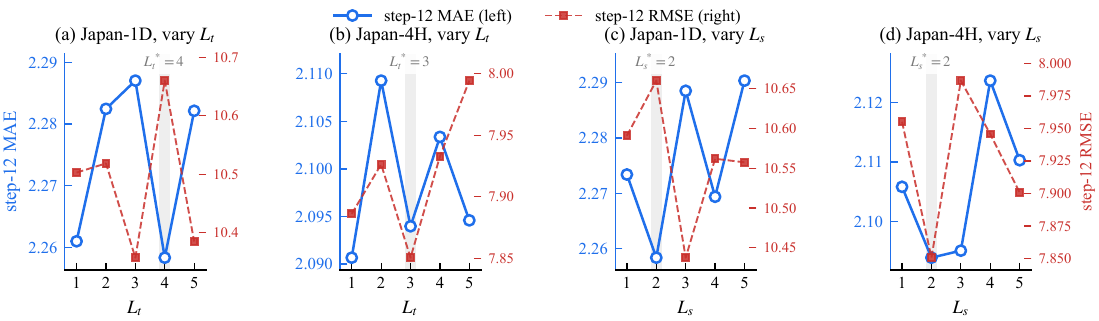}
  % \vspace{-0.7cm}
  \caption{Parameter sensitivity to the number of temporal attention layers $L_t$ and spatial attention layers $L_s$ on Japan-1D and Japan-4H. The chosen settings ($L_t{=}4$, $L_s{=}2$ on Japan-1D; $L_t{=}3$, $L_s{=}2$ on Japan-4H) lie within $0.05$ of the per-sweep MAE optimum at the 12-step horizon.}
  \label{fig:parameter_sensitivity}
  % \vspace{-0.5cm}
\end{figure}

Finally, we study the impacts of the hyper-parameters on the performance of radiation forecasting. 
We evaluate the impact of the number of temporal attention layers $L_t$ and the number of spatial attention layers $L_s$. The results on MAE, RMSE, and MAPE are reported in Figure~\ref{fig:parameter_sensitivity}. 
First, we vary $L_t$ from 1 to 5. As can be seen in Figure~\ref{fig:parameter_sensitivity}, the performance shows a fluctuation when increasing $L_t$ from 1 to 3, and drops slightly by further increasing $L_t$ from 3 to 5 on both Japan-4H and Japan-1D datasets. Overall, the optimum lies at ${L_t}{=}4$ on Japan-1D and ${L_t}{=}3$ on Japan-4H, with all sweep points within $0.05$ MAE of the per-dataset optimum, indicating that NRFormer+ is robust to this hyperparameter within the explored range.
The possible reason is that a small $L_t$ is insufficient to capture temporal correlation information, whereas too large $L_t$ may introduce redundant and noisy information for our task, leading to performance degradation.
We vary $L_s$ from 1 to 5 and observe that our model achieves optimal performance when $L_s$ is set to 2. Increasing or decreasing $L_s$ beyond this point results in a decline in performance. This is primarily because a lower $L_s$ fails to provide adequate spatio-temporal correlation of radiation propagation. 
Additionally, we noticed a performance drop when more spatial attention layers were added. 
This phenomenon might be caused by the fact that in an imbalanced spatial distribution scenario, using an excessive number of $L_s$ can lead to overfitting, particularly for nodes with highly dense connections.

\subsection{Complexity Analysis}
\label{sec:complexity}
We analyze the computational complexity of NRFormer+ with respect to the number of stations $N$, input window $P$, and hidden dimension $D$.

\textbf{Temporal attention.} The point-wise temporal self-attention operates independently across $N$ stations, yielding $\mathcal{O}(N \cdot P^2 \cdot D)$ per layer. Since $P=24$ is fixed and small, this is effectively $\mathcal{O}(N \cdot D)$.

\textbf{Physics-informed atmospheric modeling.} The atmospheric diffusion module involves MLP operations over $N$ stations, diffusion coefficient estimation is $\mathcal{O}(N \cdot (C_m + 2) \cdot 64)$, Laplacian computation is $\mathcal{O}(N)$, and physics encoding is $\mathcal{O}(N \cdot D)$. The enhanced meteorological encoder is $\mathcal{O}(N \cdot P \cdot D)$. Both are linear in $N$.

\textbf{Spatial attention.} The macro-scale spatial attention computes all-pair attention, requiring $\mathcal{O}(N^2 \cdot D)$ per layer. The proximity-constrained attention has the same worst-case complexity but is significantly faster in practice due to the sparse mask. With $L_s$ alternating layers, the total spatial complexity is $\mathcal{O}(L_s \cdot N^2 \cdot D)$.

\textbf{Overall complexity.} The dominant term is the spatial attention at $\mathcal{O}(L_s \cdot N^2 \cdot D)$, which is consistent with NRFormer. The additional physics and context modules add about $\mathcal{O}(N \cdot D^2)$ compute and $\mathcal{O}(D^2)$ parameters (from the physics encoder, meteorological encoder, temporal encoder, and cross-feature module). Since $N \gg D$, this overhead is asymptotically subdominant to the spatial term and constitutes only a modest increase over NRFormer.

\end{document}